\documentclass{article}
\usepackage[preprint]{colm2026_conference}

\usepackage{microtype}
\usepackage{hyperref}
\usepackage{url}
\usepackage{booktabs}
\usepackage{multirow}
\usepackage{graphicx}
\usepackage{amsmath}
\usepackage{amssymb}
\usepackage{xcolor}
\usepackage{colortbl}
\usepackage{pifont}
\usepackage{tikz}
\usetikzlibrary{shapes.geometric, arrows.meta, positioning, fit, backgrounds, calc}

\usepackage{lineno}

\definecolor{darkblue}{rgb}{0, 0, 0.5}
\hypersetup{colorlinks=true, citecolor=darkblue, linkcolor=darkblue, urlcolor=darkblue}

\newcommand{\bench}{\textsc{OccuBench}}

\title{\bench{}: Evaluating AI Agents on Real-World Professional Tasks via Language Environment Simulation}

\author{
Xiaomeng Hu$^{1,2*\dagger}$ \quad
Yinger Zhang$^{1*}$ \quad
Fei Huang$^{1\ddagger}$ \quad
Jianhong Tu$^{1\ddagger}$ \quad
Yang Su$^{1}$ \\
\textbf{
Lianghao Deng$^{1}$ \quad
Yuxuan Liu$^{1}$ \quad
Yantao Liu$^{1}$ \quad
Dayiheng Liu$^{1}$ \quad
Tsung-Yi Ho$^{2\ddagger}$
} \\[6pt]
$^{1}$Qwen Team, Alibaba Group \quad
$^{2}$The Chinese University of Hong Kong \\[3pt]
{\small $^{*}$Equal contribution \quad $^{\dagger}$Work done during internship at Qwen Team \quad $^{\ddagger}$Corresponding author}
}

\begin{document}

\ifcolmsubmission
\linenumbers
\fi

\maketitle

\vspace{-1.2em}
\begin{center}
\small
\hspace*{-3.0em}%
\raisebox{-0.15em}{\includegraphics[height=1em]{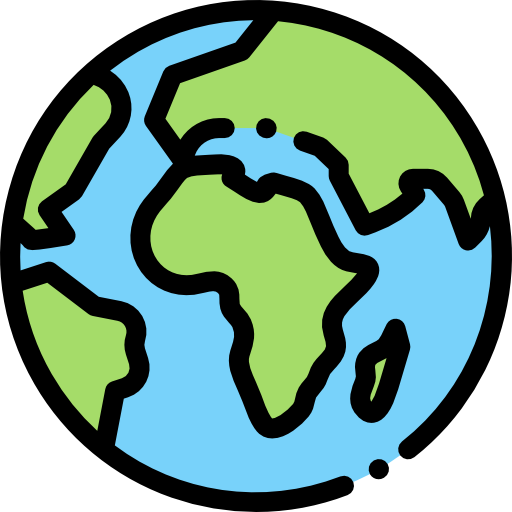}}\hspace{0.2em}\href{https://gregxmhu.github.io/OccuBench-website/}{Project Page}%
\hspace{1em}$\cdot$\hspace{1em}%
\raisebox{-0.15em}{\includegraphics[height=1em]{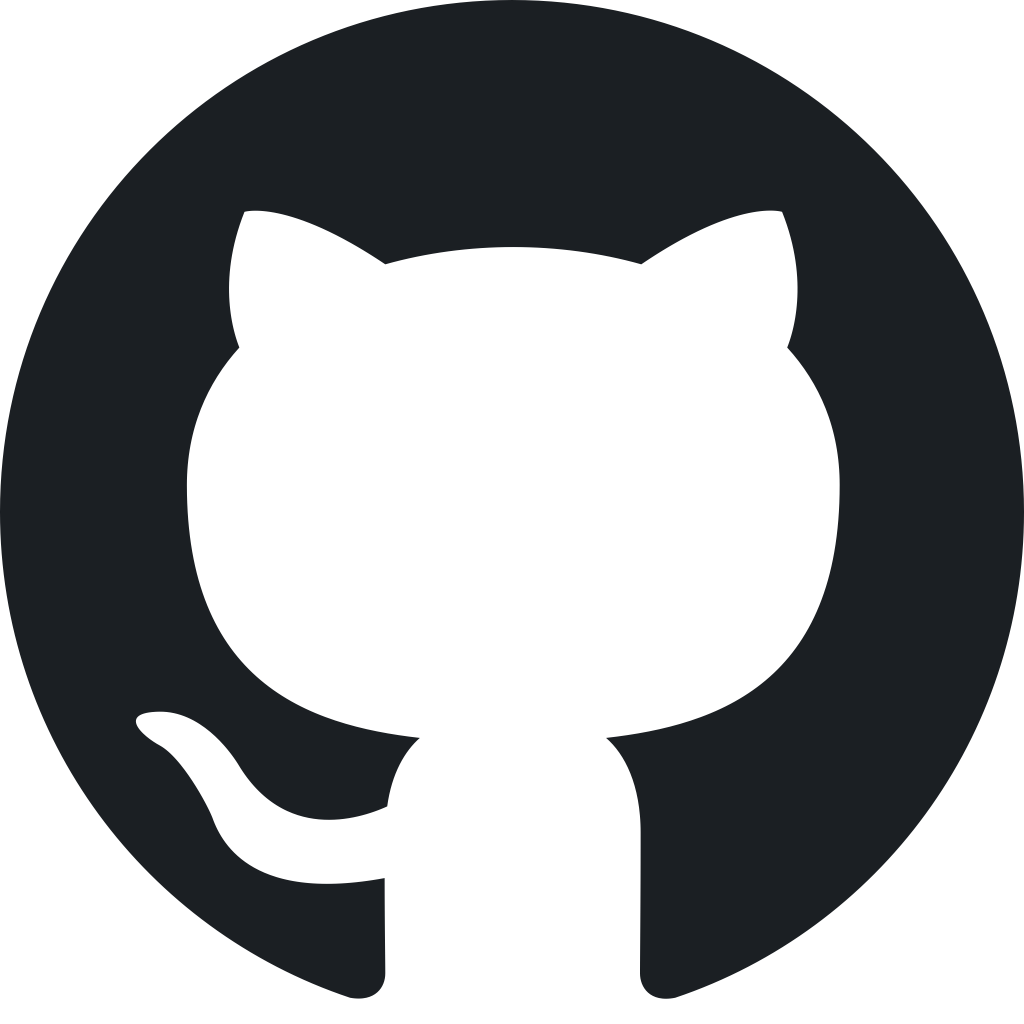}}\hspace{0.2em}\href{https://github.com/GregxmHu/OccuBench}{Code}%
\hspace{1em}$\cdot$\hspace{1em}%
\raisebox{-0.15em}{\includegraphics[height=1em]{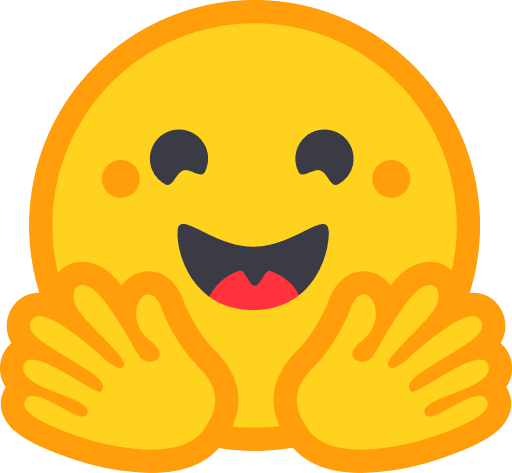}}\hspace{0.2em}\href{https://huggingface.co/datasets/gregH/OccuBench}{Data}%
\end{center}
\vspace{-0.3em}
\begin{center}
\small
Correspondence: \texttt{\{xmhu23,tyho\}@cse.cuhk.edu.hk}, \texttt{\{zhangyinger.zye,feihu.hf,tujianhong.tjh\}@alibaba-inc.com}
\end{center}
\vspace{0.5em}

\begin{abstract}
AI agents are expected to perform professional work across hundreds of occupational domains (from emergency department triage to nuclear reactor safety monitoring to customs import processing), yet existing benchmarks can only evaluate agents in the few domains where public environments exist. We introduce \bench{}, a benchmark covering \textbf{100 real-world professional task scenarios} across \textbf{10 industry categories} and \textbf{65 specialized domains}, enabled by \emph{Language Environment Simulators} (LESs) that simulate domain-specific environments through LLM-driven tool response generation. Our multi-agent synthesis pipeline automatically produces evaluation instances with guaranteed solvability, calibrated difficulty, and document-grounded diversity. \bench{} evaluates agents along two complementary dimensions: task completion across professional domains and environmental robustness under controlled fault injection (explicit errors, implicit data degradation, and mixed faults). We evaluate 15 frontier models across 8 model families and find that: (1) no single model dominates all industries, as each has a distinct occupational capability profile; (2) implicit faults (truncated data, missing fields) are harder than both explicit errors (timeouts, 500s) and mixed faults, because they lack overt error signals and require the agent to independently detect data degradation; (3) larger models, newer generations, and higher reasoning effort consistently improve performance. GPT-5.2 improves by 27.5 points from minimal to maximum reasoning effort; and (4) strong agents are not necessarily strong environment simulators. Simulator quality is critical for LES-based evaluation reliability. \bench{} provides the first systematic cross-industry evaluation of AI agents on professional occupational tasks.
\end{abstract}

\section{Introduction}

AI agents are increasingly expected to perform professional work across diverse occupational domains: triaging emergency patients, auditing financial reports, scheduling factory production lines, responding to network intrusions, processing customs declarations, and coordinating wildfire evacuations. These scenarios represent the highest-value applications of AI agent technology, where autonomous decision-making through multi-step tool use can augment or replace costly human expertise.

However, a fundamental evaluation gap exists: \textbf{the professional domains where agents would deliver the most value are precisely the domains where no benchmarks exist}. Consider the following questions that no existing benchmark can answer:

\begin{itemize}
    \item Can an agent triage patients in an emergency department? \emph{No public environment exists.}
    \item Can an agent manage a nuclear reactor safety alert? \emph{No benchmark covers this.}
    \item Can an agent process customs import declarations? \emph{No API is available.}
    \item Can an agent control greenhouse irrigation based on sensor data? \emph{No testbed exists.}
\end{itemize}

This is not a collection of edge cases; it is the \emph{default state} for the vast majority of professional work. Existing agent benchmarks are structurally unable to address these domains due to several fundamental limitations:

\paragraph{The Untestable Majority.} The professional domains where AI agents are most needed, including healthcare, finance, legal, manufacturing, energy, governance, and logistics, which are bound to enterprise systems with no public APIs, no external access, and irreversible real-world consequences. Current benchmarks are confined to domains with available environments: WebArena~\citep{webarena} to web browsing, OSWorld~\citep{osworld} to desktop operations, SWE-bench~\citep{swebench} to code repositories, and TAU-bench~\citep{taubench} to retail and airline APIs. The result is a severe evaluation blind spot covering the vast majority of high-value professional work.

\paragraph{Prohibitive Scaling Cost.} Even within covered domains, each benchmark is constrained by its environment implementation. Adding a new domain to WebArena requires deploying and configuring entire web applications; extending TAU-bench requires integrating new real APIs or manually writing simulators. This engineering overhead makes scaling to dozens or hundreds of professional domains practically infeasible.

\paragraph{No Robustness Evaluation.} Real-world environments are noisy: APIs time out, data arrives incomplete, services degrade silently. Yet existing benchmarks evaluate agents exclusively on the ``happy path,'' providing no systematic assessment of how agents handle environmental faults. This gap is critical for production deployment decisions.

\paragraph{Our Approach: Language Environment Simulator.} Our key observation is that \textbf{the environment itself can be simulated by an LLM}. Given a configuration $c = (system\_prompt, tool\_schema, initial\_state, state\_description)$, an LLM becomes a stateful, interactive environment simulator, i.e., a \emph{Language Environment Simulator} (LES). As long as an LLM understands the operational logic of a domain, it can simulate tool responses for that domain. This transforms environment construction from an engineering problem into a configuration problem, extending benchmark coverage from ``domains with public environments'' to \underline{``any domain an LLM can understand.''}

Based on LESs, we present \bench{}, a benchmark covering \textbf{100 real-world professional task scenarios} across \textbf{10 industry categories} and \textbf{65 specialized domains}, with \textbf{382 evaluation instances}. Each scenario corresponds to a real human job role (emergency triage nurse, operations engineer, customs officer, production scheduler), ensuring that evaluation results directly reflect an agent's fitness for professional work.

\bench{} evaluates agents along two complementary dimensions:
\begin{enumerate}
    \item \textbf{Task Completion}: Multi-step decision-making across 10 industry categories, revealing each model's cross-industry capability profile.
    \item \textbf{Environmental Robustness}: Performance under controlled fault injection, including explicit errors (timeouts, 500s), implicit degradation (truncated data, missing fields), and mixed faults, quantifying an agent's resilience to real-world environmental noise.
\end{enumerate}

We evaluate 15 frontier models spanning 8 model families and make the following key findings:

\begin{itemize}
    \item \textbf{No single model dominates all industries}: Gemini 3.1 Pro leads in Education (84\%) and Science (81\%) but struggles in Healthcare (62\%); Claude Opus 4.6 excels in Transportation (77\%) but trails in Commerce (53\%). Every model has blind spots invisible to single-domain benchmarks.

    \item \textbf{Implicit faults are harder than explicit faults}: Average performance under E2 (implicit, 53.4\%) drops far more than E1 (explicit, 62.6\%) from E0 (67.5\%). Implicit faults lack overt error signals and require agents to independently detect data degradation, a capability most models lack.

    \item \textbf{Scaling consistently improves performance}: Larger models outperform smaller variants within every family, newer generations outperform older ones, and higher reasoning effort yields better results. GPT-5.2 improves by 27.5 points from \texttt{none} to \texttt{xhigh} effort.

    \item \textbf{Strong agents are not necessarily strong simulators}: GPT-5.2 ranks first as an agent (79.6\%) but produces the worst environment simulation quality. When using a sufficiently capable simulator, pairwise ranking agreement reaches 85.7\%, confirming that LES-based evaluation is reliable.
\end{itemize}

\section{Related Work}

\paragraph{Agent Benchmarks.}
Agent benchmarks can be categorized by environment type. \emph{Web environments}: WebArena~\citep{webarena} deploys real websites for browser-based tasks; VisualWebArena~\citep{visualwebarena} and Mind2Web~\citep{mind2web} extend to visual and cross-domain web interaction; WorkArena~\citep{workarena} targets enterprise knowledge work on ServiceNow; BrowseComp~\citep{browsecomp} evaluates deep web navigation. \emph{OS and mobile environments}: OSWorld~\citep{osworld} provides full operating system virtual machines; AndroidWorld~\citep{androidworld} benchmarks mobile app automation; MobileBench~\citep{mobilebench} evaluates LLM-based mobile agents. \emph{Code environments}: SWE-bench~\citep{swebench} evaluates repository-level issue resolution; InterCode~\citep{intercode} provides interactive coding with execution feedback; Terminal-Bench~\citep{terminalbench} tests agents in real terminal environments. \emph{Tool and API environments}: TAU-bench~\citep{taubench} evaluates tool-agent-user interaction; BFCL~\citep{bfcl} benchmarks function calling; AgentBench~\citep{agentbench} covers 8 distinct environments; ToolLLM~\citep{toolbench} evaluates across 16,000+ real-world APIs; GAIA~\citep{gaia} tests general assistant capabilities with multi-modal reasoning and tool use; MINT~\citep{mint} evaluates multi-turn interaction with tools; and MCP-Bench~\citep{mcpbench}, MCP-Atlas~\citep{mcpatlas}, MCPMark~\citep{mcpmark}, and Toolathlon~\citep{toolathlon} benchmark tool-use competency through real MCP servers.

All share fundamental limitations: (1) environments require substantial engineering to construct and maintain; (2) test sets are static and vulnerable to data contamination; (3) no systematic environmental robustness evaluation; and most critically, (4) \textbf{domain coverage is extremely limited}: all existing benchmarks combined cover only web browsing, code editing, desktop operations, and a handful of API domains, leaving the vast majority of professional occupational tasks untestable.

\paragraph{Real-World Professional Task Evaluation.}
Several recent benchmarks target economically valuable professional work. GDPVal~\citep{gdpval} covers 44 occupations across 9 industries with 1,320 tasks graded by industry experts, focusing on \emph{output-quality} tasks (writing legal briefs, creating presentations). \$OneMillion-Bench~\citep{onemillionbench} evaluates 400 expert-curated tasks across Law, Finance, Industry, Healthcare, and Natural Science, where each task is assigned a monetary value based on senior professional hourly rates. TheAgentCompany~\citep{theagentcompany} evaluates agents as digital workers performing consequential real-world tasks. SWE-Lancer~\citep{swelancer} maps agent performance to monetary value through 1,400+ real freelance software engineering tasks worth \$1M total. Claw-Eval~\citep{claweval} introduces an end-to-end evaluation suite of 300 human-verified tasks spanning 9 categories with trajectory-aware grading over 2,159 fine-grained rubric items, evaluating completion, safety, and robustness of autonomous agents. These benchmarks are complementary to \bench{}: GDPVal and \$OneMillion-Bench measure deliverable quality via rubric-based grading, TheAgentCompany and SWE-Lancer focus on software-adjacent work, while \bench{} measures \emph{interactive decision-making} across 65 specialized domains, from emergency triage to nuclear reactor monitoring, requiring multi-step tool use, state tracking, and error handling in stateful environments.

\paragraph{Context Learning.}
CL-bench~\citep{clbench} evaluates models' ability to learn from task-specific context containing new knowledge beyond pre-training, covering 500 complex contexts with 1,899 tasks. While CL-bench tests context-dependent reasoning, \bench{} tests context-dependent \emph{action}: agents must not only understand domain-specific contexts but execute multi-step tool-use workflows within them, handling environmental feedback and adapting to unexpected conditions.

\paragraph{World Models and Environment Simulation.}
Traditional world models such as Dreamer~\citep{dreamer} and IRIS~\citep{iris} learn environment dynamics from data but are limited to low-dimensional state spaces. LLM-based simulation approaches like Generative Agents~\citep{generativeagents} use language models to simulate social behavior but do not involve tool-use interaction or stateful task execution. Recent work has explored LLMs as environment simulators more directly: \citet{simenv} train reasoning models to simulate environments for agent training; \citet{webworldmodel} show that LLMs can serve as world models of the internet for web agent planning; WebWorld~\citep{webworld} trains the first open-web simulator at scale on 1M+ interactions, demonstrating that world models can enable effective agent training and inference-time search; ViMo~\citep{vimo} builds generative visual world models for GUI agents; and self-play approaches~\citep{selfplaywm} internalize world models for agentic reinforcement learning. Our Language Environment Simulator approach occupies a distinct niche: using LLMs to simulate \emph{tool-response-level} environment interaction for \emph{evaluation} rather than training, supporting stateful multi-step professional tasks with realistic action spaces across 100 scenarios and 65 specialized domains.

\section{Language Environment Simulator}

\subsection{Formalization}

We define a Language Environment Simulator (LES) as a function:
\begin{equation}
    (s_{t+1}, o_{t+1}) = f_\theta(s_t, a_t; c)
\end{equation}
where $c = (system\_prompt, tool\_schema, initial\_state, state\_description)$ is the environment configuration, $s_t$ is the latent environment state maintained implicitly by the LLM through its context window, $a_t$ is the agent's action (a tool call with name and arguments), and $o_{t+1}$ is the observation returned to the agent (a structured JSON tool response).

Unlike traditional world models that learn state transition functions from data, LESs leverage the LLM's pre-trained knowledge of domain-specific operational logic. The system prompt encodes the simulation rules; the tool schema defines the action space; the initial state and state description constrain the LLM to maintain causal consistency across interactions.

\subsection{Environment Configuration}

Each LES environment is fully specified by four components:

\begin{itemize}
    \item \textbf{System Prompt}: Defines the environment's behavioral rules, simulation logic, error handling protocols, and output format constraints. For example, a hotel revenue management environment's system prompt specifies pricing rules, occupancy calculations, and the relationship between ADR, CPOR, and revenue metrics.

    \item \textbf{Tool Schema}: Defines the agent's action space as a set of callable functions with typed parameters and example outputs. Each environment contains 2--10 tools (median 5) reflecting realistic operational interfaces.

    \item \textbf{Initial State}: A structured JSON object specifying the environment's starting conditions (e.g., room inventory, patient queue, network topology).

    \item \textbf{State Description}: Semantic annotations for each state field, guiding the LLM to maintain causal consistency (e.g., ``remaining\_inventory decreases after each booking'').
\end{itemize}

\subsection{Why LLMs Can Serve as Language Environment Simulators}

LLMs are effective language environment simulators for professional tasks because: (1) \textbf{Format priors}: Pre-training on vast API documentation and tool-call logs provides strong priors for generating well-formatted tool responses. (2) \textbf{Domain knowledge}: LLMs encode operational logic for hundreds of professional domains, from hospital triage protocols to network firewall rules. (3) \textbf{State maintenance}: The combination of system prompt constraints and in-context state tracking enables coherent multi-turn simulation. (4) \textbf{Edge case handling}: LLMs handle unexpected inputs more gracefully than rule-based simulators, generating reasonable error responses for out-of-bounds parameters.

Figure~\ref{fig:lwm_interaction} illustrates the interaction loop between the agent and the LES at evaluation time.

\begin{figure}[t]
\centering
\resizebox{\linewidth}{!}{%
\begin{tikzpicture}[
    >=Stealth,
    every node/.style={font=\normalsize},
    box/.style={draw, rounded corners=5pt, minimum height=1.6cm, minimum width=3.2cm, align=center, line width=0.8pt},
    config/.style={draw, rounded corners=3pt, minimum height=0.6cm, font=\small, align=center},
    groupbox/.style={draw, rounded corners=6pt, dashed, inner sep=8pt, line width=0.6pt},
    arr/.style={->, thick, color=black!70},
    dataarr/.style={->, thick, color=blue!60!black},
]

\node[box, fill=blue!8] (lwm) {\textbf{Language Env.}\\ \textbf{Simulator (LES)}};

\node[font=\normalsize, above=0.4cm of lwm] (formula) {$o_t = \mathrm{LES}(a_t, c,\; H_{t-1})$};

\node[box, fill=orange!12, above left=1.6cm and 3cm of lwm] (agent) {\textbf{Agent LLM}\\{\small(model under test)}};

\node[box, fill=green!8, above right=1.6cm and 3cm of lwm] (verifier) {\textbf{Verifier}\\{\small(rubric-based)}};

\node[draw, rounded corners=4pt, fill=yellow!12, minimum width=3.2cm, minimum height=1.4cm, font=\small, align=center, below left=1.6cm and 3cm of lwm] (hist) {\textbf{History} $H_{t-1}$\\[3pt]$a_1, o_1, \ldots,$\\$a_{t-1}, o_{t-1}$};

\node[config, fill=white, below right=1.2cm and 2.2cm of lwm] (sp) {System Prompt};
\node[config, fill=white, below=0.15cm of sp] (tools) {Tool Schema};
\node[config, fill=white, below=0.15cm of tools] (init) {Initial State};
\node[config, fill=white, below=0.15cm of init] (sd) {State Description};

\begin{scope}[on background layer]
\node[groupbox, fill=gray!6, fit=(sp)(tools)(init)(sd), label={[font=\small\bfseries, fill=white, inner sep=2pt]above:Configuration $c$}] (configgroup) {};
\end{scope}

\draw[dataarr, line width=1.2pt] ([yshift=0.12cm]agent.south east) -- node[above, font=\small, sloped] {tool call $a_t$} ([yshift=0.12cm]lwm.north west);
\draw[dataarr, line width=1.2pt] ([yshift=-0.12cm]lwm.north west) -- node[below, font=\small, sloped] {observation $o_t$} ([yshift=-0.12cm]agent.south east);

\draw[arr, line width=1.2pt] (lwm.north east) -- node[above, font=\small, sloped] {trajectory} (verifier.south west);
\node[font=\small, right=0.15cm of verifier] {$\rightarrow$ pass/fail};

\draw[arr, dashed, blue!40!black, line width=1.2pt] (hist.north east) -- (lwm.south west);

\draw[arr, dashed, gray!60, line width=1.2pt] (configgroup.north west) -- (lwm.south east);

\end{tikzpicture}%
}
\caption{LES evaluation loop. At each step, the agent issues a tool call $a_t$; the LES generates an observation $o_t$ conditioned on its configuration $c$ and conversation history $H_{t-1}$. State is maintained implicitly through in-context history. The trajectory is scored by a rubric-based verifier.}
\label{fig:lwm_interaction}
\end{figure}

\section{Multi-Agent Synthesis Pipeline}

Each evaluation instance must satisfy four conditions: (1) \textbf{solvable}: a valid solution exists and is verified; (2) \textbf{verifiable}: clear, automated success criteria; (3) \textbf{discriminative}: calibrated difficulty that distinguishes agent capabilities; and (4) \textbf{diverse}: structural variation across instances.

To ensure diversity, we design 16 non-overlapping sub-topics per scenario and construct a professional reference document for each, covering domain terminology, workflows, state variables, edge cases, and constraints. These documents ground all subsequent generation, ensuring instances differ structurally rather than superficially.

We employ a multi-agent synthesis pipeline powered by Gemini-3-Flash-Preview as the Language Environment Simulator. The pipeline generates environment configurations, task instructions, tool definitions, solution plans, and verification rubrics. Each task is executed multiple times with and without a reference plan to verify solvability and calibrate difficulty. A majority-vote verifier assesses trajectories against rubrics, and a repair module diagnoses and fixes failures before re-execution. Tasks that are trivially easy (100\% autonomous success), unsolvable (0\% success), or have invalid tool schemas are filtered out.

\section{\bench{} Benchmark}
\label{sec:benchmark}

\subsection{Scale and Coverage}

\bench{} covers 100 professional task scenarios across 10 industry categories and 65 specialized domains (Table~\ref{tab:categories}). Each scenario maps to a real human job role, ensuring evaluation results have direct practical relevance. After synthesis and quality filtering, removing instances where all difficulty levels are trivially solved (100\% autonomous success rate), instances that are unsolvable (0\% success rate), and instances with invalid tool schemas. Our evaluation set contains \textbf{382 solvable task instances} spanning all 100 scenarios. For each task, we select the difficulty level with the lowest autonomous success rate to maximize discriminative power. The final dataset averages 5.5 tools and 16.2 tool calls per task.

\begin{table}[t]
\centering
\small
\caption{Industry categories and representative scenarios in \bench{}.}
\label{tab:categories}
\begin{tabular}{lrl}
\toprule
\textbf{Category} & \textbf{\#} & \textbf{Representative Scenarios} \\
\midrule
Business \& Enterprise & 19 & Resume screening, expense auditing, AML review \\
Technology \& IT & 16 & Linux ops, CI/CD recovery, intrusion response \\
Industrial \& Engineering & 12 & Production scheduling, mine ventilation \\
Transportation \& Logistics & 11 & Last-mile delivery, train dispatch \\
Commerce \& Consumer & 9 & Dynamic pricing, hotel revenue mgmt. \\
Education \& Culture & 8 & Adaptive curriculum, fact-checking \\
Healthcare \& Life Sciences & 7 & Emergency triage, drug interaction screening \\
Public Service \& Governance & 7 & Permit processing, wildfire evacuation \\
Agriculture \& Environment & 7 & Irrigation control, crop disease diagnosis \\
Science \& Research & 4 & Telescope scheduling, excavation planning \\
\bottomrule
\end{tabular}
\end{table}

\paragraph{Scenario Design Principles.} (1) \emph{Real job mapping}: Each scenario corresponds to an actual professional role, not an abstract task. (2) \emph{Domain balance}: No single domain contributes more than 3 scenarios. (3) \emph{Irreplaceability}: The majority of scenarios (nuclear safety, drug screening, emergency coordination) are untestable by any existing benchmark. (4) \emph{Multi-step interaction}: All scenarios require multi-turn state transitions, not single-step function calls.

\subsection{Environmental Fault Injection}

\bench{} evaluates agent robustness through controlled fault injection at evaluation time. All data is synthesized in clean environments (E0); faults are injected by appending fault rules to the LES's system prompt during evaluation.

\textbf{E0 (Clean):} No faults. Baseline performance.

\textbf{E1 (Explicit Faults):} The LES randomly injects clearly visible error responses: HTTP 500 Internal Server Error, TimeoutError, ConnectionRefused, ServiceUnavailable. These faults have \emph{clear error signals}: the agent knows the call failed. The correct behavior is to retry.

\textbf{E2 (Implicit Faults):} The LES returns degraded responses with \emph{no error signal}: truncated data (missing fields), incomplete lists (only first 1--2 items), empty/null fields, or stale cached values. The response appears superficially correct. The correct behavior is to detect the quality issue and re-query.

\textbf{E3 (Mixed):} Approximately half explicit, half implicit faults.

All faults are \emph{transient} (retrying recovers normal results), \emph{spaced} across the interaction (not concentrated at the start), and parameterized by two independent controls: \texttt{fault\_count} (number of fault events, default 2) and \texttt{fault\_duration} (consecutive tool calls affected per event, default 2).

\subsection{Evaluation Metrics}

\textbf{Completion Rate (CR):} Fraction of 382 tasks where the agent's trajectory passes automated verification against the rubric. We report all rates over the full 382-task denominator.

\textbf{Robustness Score (R):} $R = \min(\text{CR}_{E1}, \text{CR}_{E2}, \text{CR}_{E3}) / \text{CR}_{E0}$, measuring worst-case resilience across all fault types. A score of 1.0 indicates no degradation; lower scores indicate greater sensitivity to environmental noise. We use the minimum.

\section{Experiments}
\label{sec:experiments}

We evaluate 15 frontier models spanning 8 model families: OpenAI (GPT-5.2~\citep{gpt5}), Anthropic (Claude Opus/Sonnet 4, 4.5, 4.6~\citep{claude4,claude45,claude46,claude46opus}), Google (Gemini 3.1 Pro, Flash-Lite~\citep{gemini3}), DeepSeek (V3.2~\citep{deepseekv32}), Moonshot (Kimi K2.5~\citep{kimik25}), MiniMax (M2.7~\citep{minimaxm27}), Zhipu (GLM-5~\citep{glm5}), and Alibaba (Qwen 3.5 Plus, Flash~\citep{qwen35}). All models use thinking/reasoning mode where available, with the default Language Environment Simulator being Gemini-3-Flash-Preview.

\subsection{Main Results: Cross-Industry Evaluation (E0)}

Table~\ref{tab:e0_category} presents completion rates across 10 industry categories for all 15 models.

\begin{table*}[t]
\centering
\small
\caption{E0 completion rate (\%) by industry category for all 15 models. All models use thinking mode; for models with adjustable reasoning effort, we set it to \texttt{high}. \textbf{Bold}: best in each category. Models sorted by average score.}
\label{tab:e0_category}
\setlength{\tabcolsep}{3.5pt}
\begin{tabular}{l|c|cccccccccc}
\toprule
\textbf{Model} & \textbf{Avg} & \textbf{Agri} & \textbf{Biz} & \textbf{Comm} & \textbf{Edu} & \textbf{Hlth} & \textbf{Ind} & \textbf{Pub} & \textbf{Sci} & \textbf{Tech} & \textbf{Trans} \\
\midrule
GPT-5.2 & 79.6 & \textbf{84} & \textbf{86} & 67 & 77 & 76 & \textbf{85} & \textbf{84} & \textbf{94} & \textbf{80} & 72 \\
Gemini 3.1 Pro & 72.3 & 68 & 73 & 75 & \textbf{84} & 62 & 73 & 72 & 81 & 78 & 60 \\
Claude Opus 4.6 & 71.5 & 74 & 78 & 53 & 75 & 76 & 73 & 68 & 62 & 68 & \textbf{77} \\
Qwen 3.5 Plus & 69.9 & 77 & 70 & \textbf{81} & 56 & \textbf{81} & 71 & 76 & 69 & 74 & 55 \\
DeepSeek V3.2 & 69.6 & 65 & 78 & 67 & 66 & 71 & 69 & 72 & 62 & 74 & 64 \\
Claude Opus 4.5 & 65.2 & 58 & 76 & 56 & 62 & 52 & 65 & 72 & 56 & 68 & 66 \\
Claude Sonnet 4.5 & 64.9 & 65 & 70 & 69 & 50 & 71 & 71 & 60 & 44 & 68 & 62 \\
Claude Sonnet 4.6 & 64.4 & 58 & 71 & 64 & 69 & 67 & 64 & 64 & 69 & 64 & 57 \\
Kimi K2.5 & 64.1 & 68 & 62 & 56 & 62 & 81 & 62 & 72 & 56 & 74 & 57 \\
GLM-5 & 62.6 & 55 & 75 & 67 & 53 & 57 & 56 & 68 & 62 & 70 & 55 \\
Claude Opus 4 & 61.3 & 52 & 75 & 50 & 53 & 57 & 58 & 76 & 81 & 66 & 51 \\
Gemini 3.1 FL & 61.3 & 68 & 70 & 58 & 53 & 67 & 58 & 68 & 62 & 68 & 45 \\
Qwen 3.5 Flash & 59.7 & 61 & 60 & 67 & 53 & 76 & 53 & 68 & 69 & 60 & 51 \\
MiniMax M2.7 & 53.9 & 48 & 60 & 56 & 31 & 57 & 60 & 60 & 62 & 64 & 40 \\
Claude Sonnet 4 & 53.4 & 35 & 63 & 61 & 38 & 57 & 51 & 76 & 31 & 60 & 47 \\
\bottomrule
\end{tabular}
\end{table*}

\paragraph{No single model dominates all industries.} GPT-5.2 leads overall (79.6\%) with the highest scores in Agriculture (84\%), Business (86\%), Industrial (85\%), and Science (94\%), but its Commerce score (67\%) is far below Qwen 3.5 Plus (81\%). Gemini 3.1 Pro ranks second (72.3\%) with the highest score in Education (84\%). Claude Opus 4.6, ranked third (71.5\%), shows the opposite pattern: strongest in Transportation (77\%) and Business (78\%) but weakest in Commerce (53\%). Qwen 3.5 Plus leads Healthcare and Commerce (both 81\%) but trails in Education (56\%).

\paragraph{Open-source models are highly competitive.} Qwen 3.5 Plus (69.9\%) and DeepSeek V3.2 (69.6\%) rank 4th and 5th, outperforming most Claude variants. This challenges the conventional assumption that closed-source models uniformly outperform open-source alternatives on professional tasks.

\begin{figure}[t]
    \centering
    \includegraphics[width=0.85\linewidth]{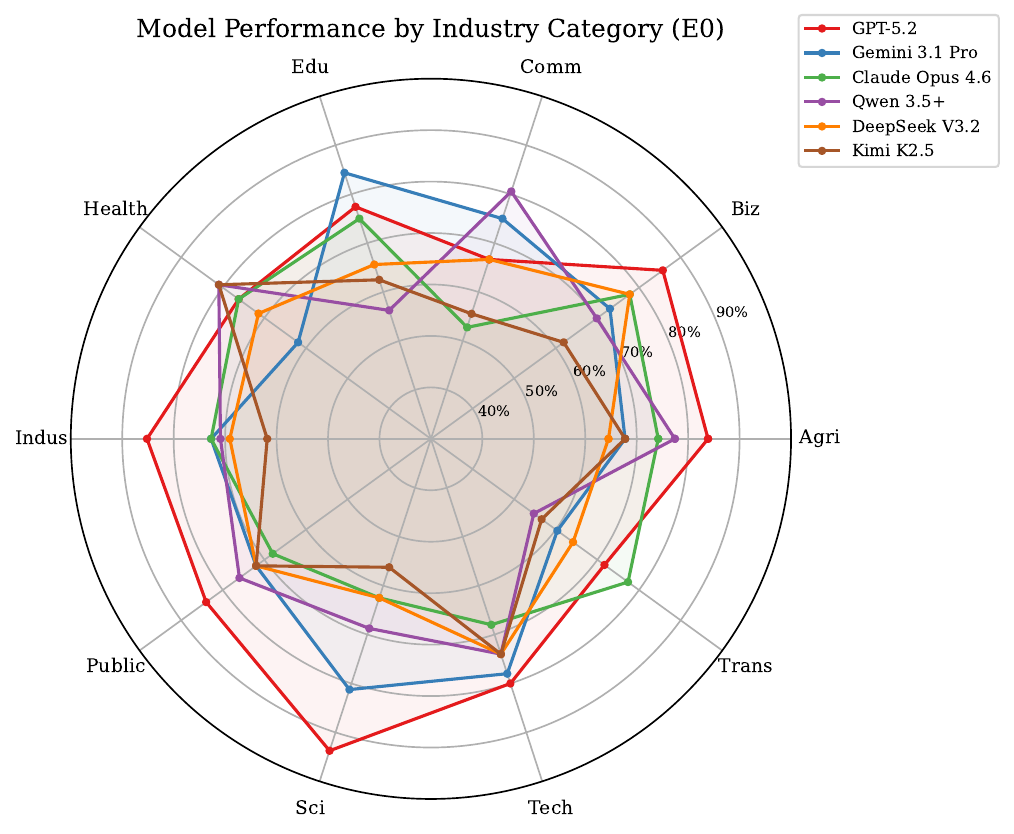}
    \caption{Radar chart showing model performance profiles across 10 industry categories (E0). Each model has a distinct shape, indicating different occupational specializations.}
    \label{fig:radar}
\end{figure}

\paragraph{Each model has a distinct occupational capability profile.} Figure~\ref{fig:radar} visualizes the top 6 models' performance across industries, revealing strikingly different capability shapes. This diversity is uniquely revealed by \bench{}'s cross-industry design and invisible to single-domain benchmarks.

\subsection{Environmental Robustness}

\begin{figure}[t]
    \centering
    \includegraphics[width=\linewidth]{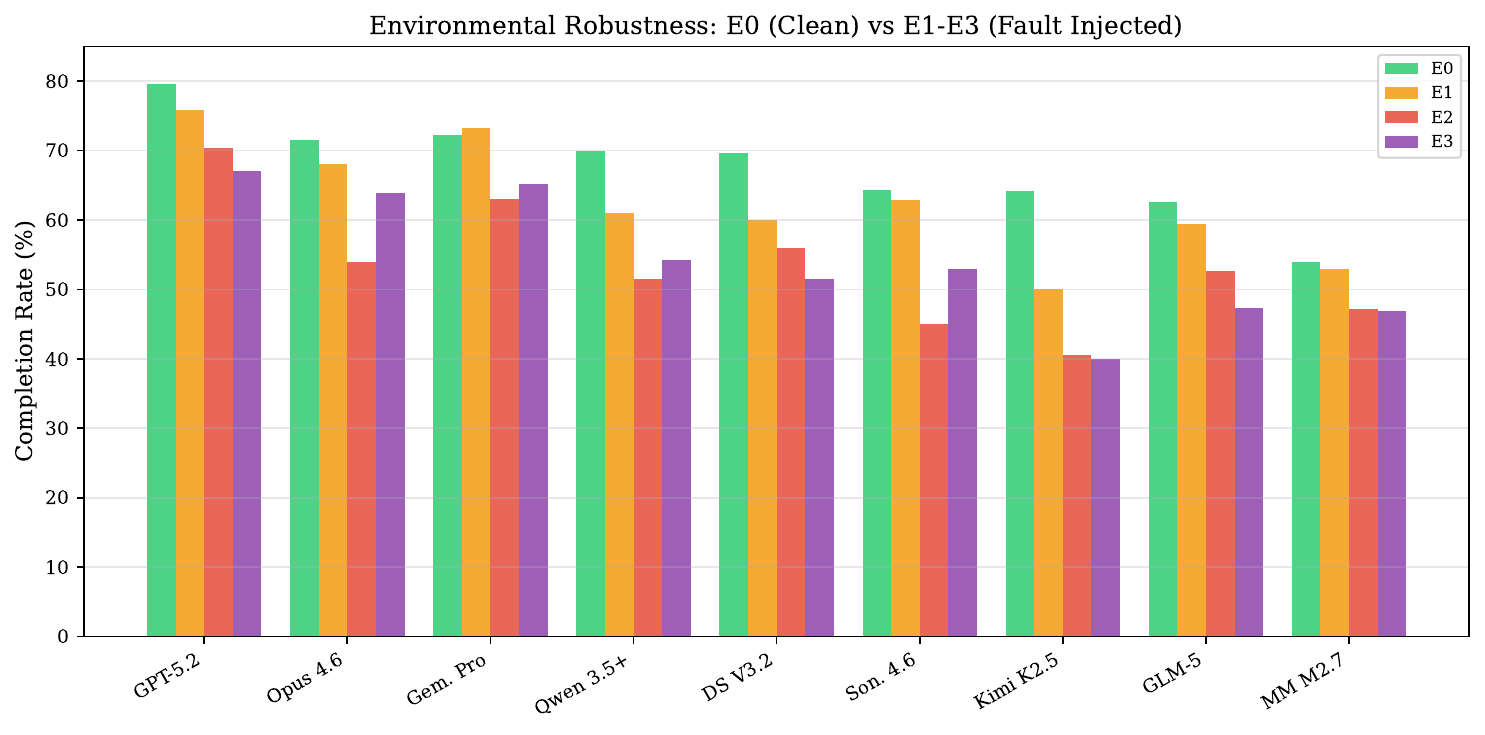}
    \caption{Completion rates under clean (E0) and fault-injected (E1--E3) environments. Models are sorted by E0 performance.}
    \label{fig:robustness}
\end{figure}

Table~\ref{tab:robustness} presents completion rates under fault injection for 9 flagship models (one per family), and Figure~\ref{fig:robustness} visualizes the comparison.

\begin{table}[t]
\centering
\small
\caption{Environmental robustness evaluation for 9 flagship models. CR = Completion Rate (\%). Rob. = Robustness ($\min(\text{CR}_{E1}, \text{CR}_{E2}, \text{CR}_{E3})/\text{CR}_{E0}$). \textbf{Bold}: best in each column. Models sorted by Robustness.}
\label{tab:robustness}
\begin{tabular}{l|cccc|c}
\toprule
\textbf{Model} & \textbf{E0} & \textbf{E1} & \textbf{E2} & \textbf{E3} & \textbf{Rob.} \\
\midrule
Gemini 3.1 Pro & 72.3 & 73.3 & 63.1 & 65.2 & \textbf{0.87} \\
MiniMax M2.7 & 53.9 & 52.9 & 47.1 & 46.9 & \textbf{0.87} \\
GPT-5.2 & \textbf{79.6} & \textbf{75.9} & \textbf{70.4} & \textbf{67.0} & 0.84 \\
GLM-5 & 62.6 & 59.4 & 52.6 & 47.4 & 0.76 \\
Claude Opus 4.6 & 71.5 & 68.1 & 53.9 & 63.9 & 0.75 \\
DeepSeek V3.2 & 69.6 & 59.9 & 56.0 & 51.6 & 0.74 \\
Qwen 3.5 Plus & 69.9 & 61.0 & 51.6 & 54.2 & 0.74 \\
Claude Sonnet 4.6 & 64.4 & 62.8 & 45.0 & 52.9 & 0.70 \\
Kimi K2.5 & 64.1 & 50.0 & 40.6 & 40.1 & 0.63 \\
\midrule
\textbf{Avg} & 67.5 & 62.6 & 53.4 & 54.4 & 0.77 \\
\bottomrule
\end{tabular}
\end{table}

\paragraph{Current agents struggle under adverse environments.} Even with only 2 fault events of 2 rounds each, performance drops substantially across the board: the average completion rate falls from 67.5\% (E0) to 53.4\% (E2), a 14.1-point decline. Even the strongest models are not immune: Claude Opus 4.6 drops 17.6 points under implicit faults (71.5\% $\rightarrow$ 53.9\%), and Qwen 3.5 Plus drops 18.3 points (69.9\% $\rightarrow$ 51.6\%). This reveals a significant gap between clean-environment capability and real-world deployment readiness.

\paragraph{Implicit faults (E2) are harder than both explicit (E1) and mixed (E3) faults.} Counter-intuitively, 4 out of 9 models perform \emph{worse} under E2 than E3, and the average E2 score (53.4\%) is lower than both E1 (62.6\%) and E3 (54.4\%). Explicit errors (timeouts, HTTP 500) provide unambiguous failure signals that prompt retry, while implicit faults (truncated data, missing fields) require the agent to independently assess response quality, a capability most models lack. E3's explicit error component partially compensates for its implicit component, making pure implicit faults (E2) the hardest environment overall.

\begin{figure}[t]
    \centering
    \includegraphics[width=\linewidth]{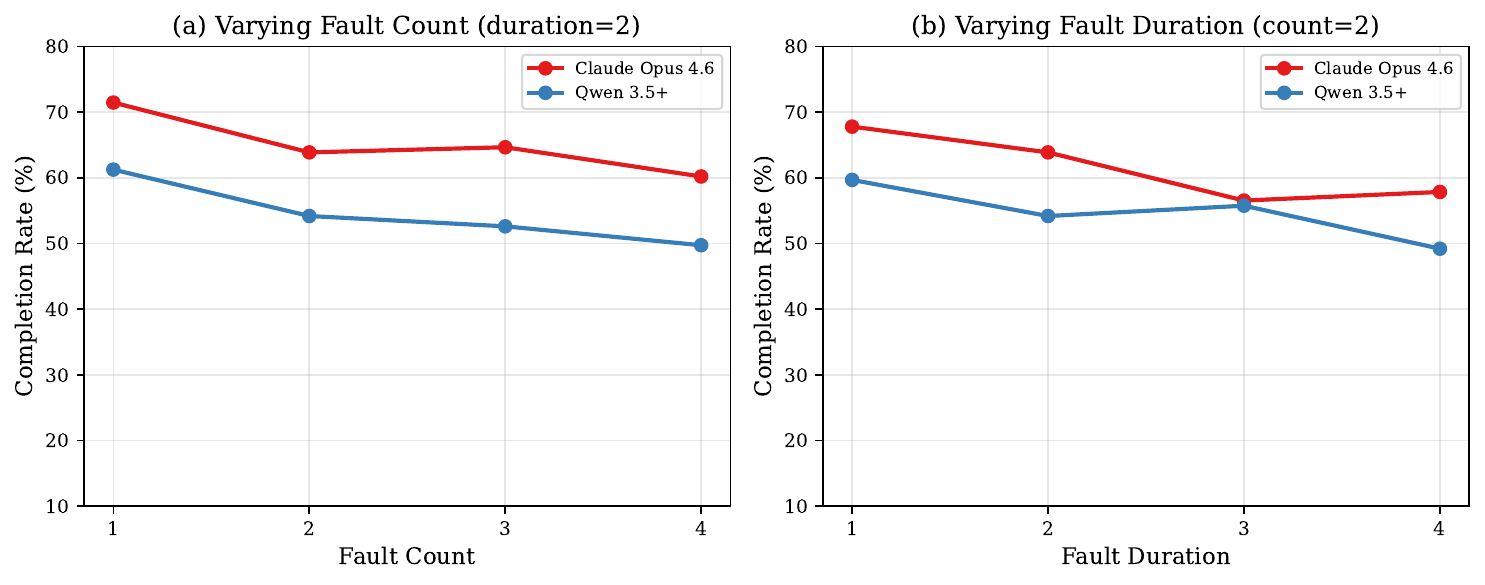}
    \caption{Fault parameter ablation under E3 mixed faults. (a) Varying fault count with fixed duration=2. (b) Varying fault duration with fixed count=2.}
    \label{fig:fault_ablation}
\end{figure}

\paragraph{Increasing fault severity deepens the challenge.} We ablate fault parameters on three representative models (Figure~\ref{fig:fault_ablation}). As fault count and duration increase beyond the default (fc=2, fd=2), performance continues to decline: Claude Opus 4.6 drops from 71.5\% (fc=1) to 60.2\% (fc=4), and from 67.8\% (fd=1) to 57.9\% (fd=4). Qwen 3.5 Plus degrades from 61.3\% to 49.7\% (count) and 59.7\% to 49.2\% (duration). These results highlight an increasingly severe challenge for deploying agents in real-world environments, where faults are not only inevitable but may be frequent and persistent.

\subsection{Model Scaling Analysis}

\begin{figure}[t]
    \centering
    \includegraphics[width=0.85\linewidth]{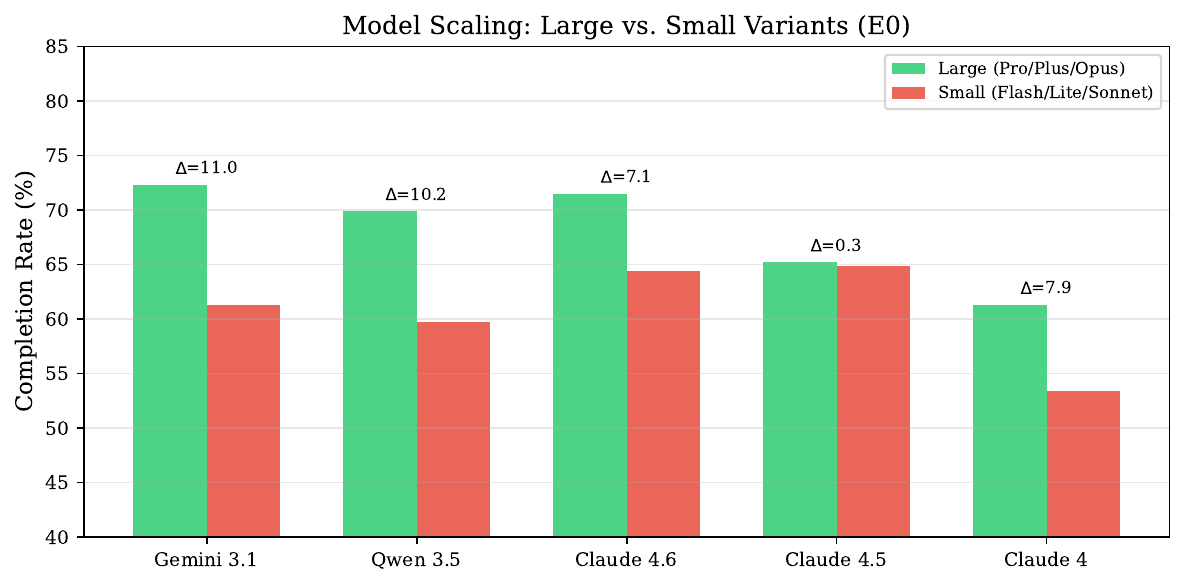}
    \caption{Large vs. small model variants within each family (E0). Gaps range from 0.3\% to 11.0\%.}
    \label{fig:scaling}
\end{figure}

\bench{} enables direct within-family comparisons between model sizes (Figure~\ref{fig:scaling}).

Larger models consistently outperform smaller counterparts, with gaps of 11.0\% (Gemini Pro vs. Flash-Lite), 10.2\% (Qwen Plus vs. Flash), and 7.1\% (Claude Opus vs. Sonnet 4.6). The notable exception is Claude 4.5, where Opus and Sonnet perform nearly identically (65.2\% vs. 64.9\%), suggesting that the 4.5 generation's architectural improvements benefited both model sizes equally.

\subsection{Generational Progress}

\begin{figure}[t]
    \centering
    \includegraphics[width=0.7\linewidth]{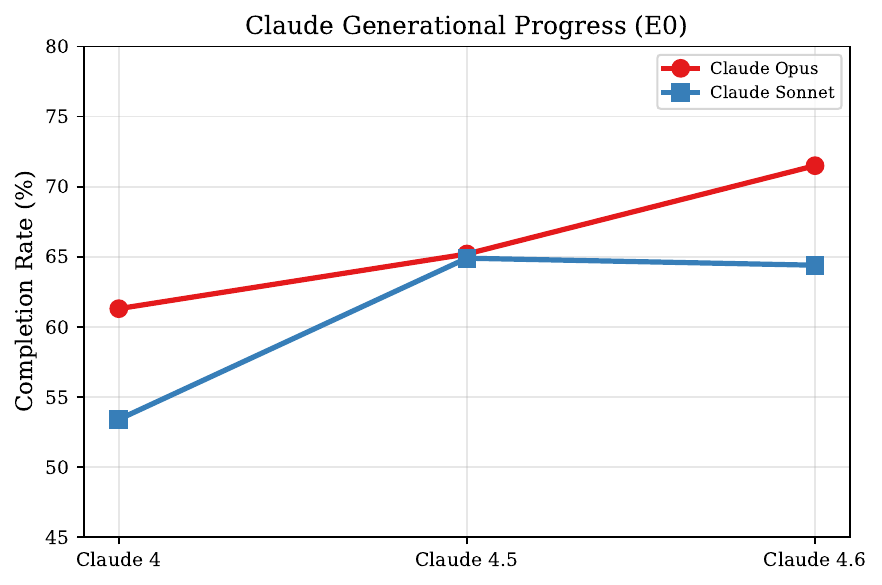}
    \caption{Claude generational progress from v4 to v4.6 (E0). Opus shows consistent improvement (+10.2\%); Sonnet shows large initial gains but slight regression from v4.5 to v4.6.}
    \label{fig:generational}
\end{figure}

Figure~\ref{fig:generational} tracks Claude's performance evolution across three generations.

Claude Opus shows consistent generational improvement: 61.3\% $\rightarrow$ 65.2\% $\rightarrow$ 71.5\% (+10.2\% total). Sonnet shows a large jump from v4 to v4.5 (+11.5\%) but slight regression from v4.5 to v4.6 ($-$0.5\%), possibly reflecting a trade-off between reasoning depth and execution efficiency in the 4.6 adaptive thinking architecture.

\subsection{Reasoning Effort Ablation}

\begin{figure}[t]
    \centering
    \includegraphics[width=0.85\linewidth]{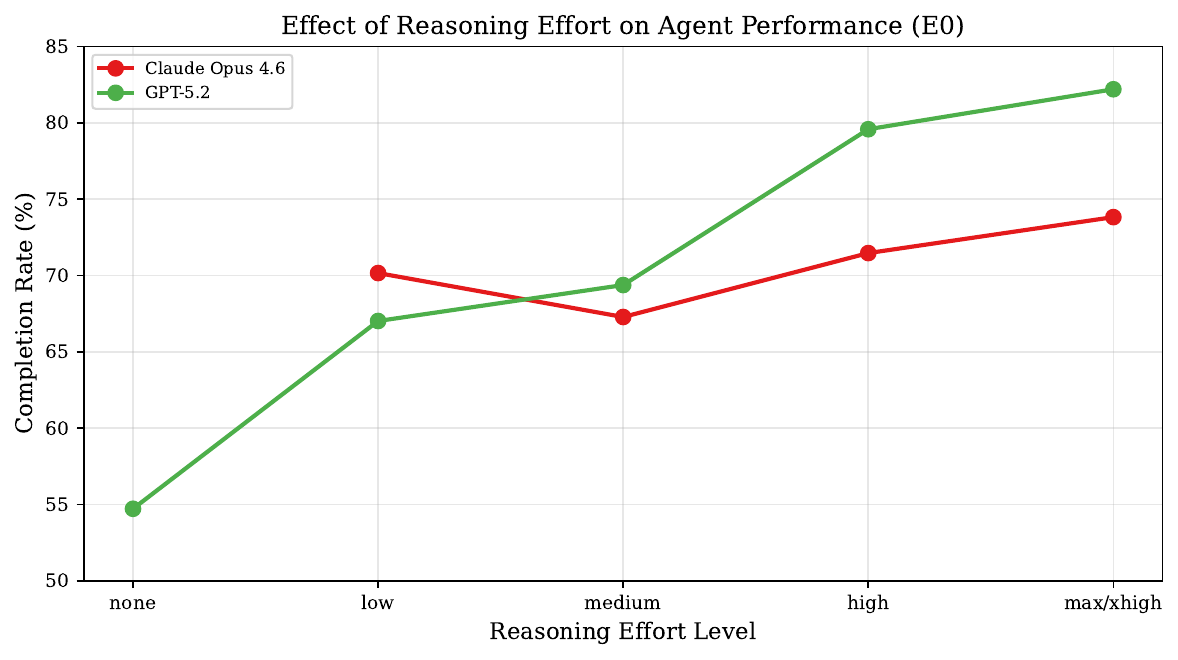}
    \caption{Effect of reasoning effort on agent performance (E0).}
    \label{fig:effort}
\end{figure}

We evaluate the effect of reasoning effort (thinking depth) on two models that support configurable effort levels: Claude Opus 4.6 and GPT-5.2 (Figure~\ref{fig:effort}).

Higher reasoning effort generally leads to better agent performance. GPT-5.2 exhibits a clear monotonic trend: scaling from \texttt{none} (54.7\%) to \texttt{xhigh} (82.2\%), a 27.5-point improvement, demonstrating that deeper reasoning directly translates to better task execution. Claude Opus 4.6 shows a similar overall trend, with its highest effort level \texttt{max} (73.8\%) outperforming \texttt{low} (70.2\%) by 3.6 points. These results suggest that allocating more compute to reasoning at inference time is a reliable strategy for improving agent performance on complex professional tasks.

\subsection{Simulator Quality Matters}

A key question for LES-based evaluation is whether a strong agent model is also a strong environment simulator. We evaluate 8 agents under three simulators: the default Gemini-3-Flash-Preview, Qwen 3.5 Plus, and GPT-5.2. Table~\ref{tab:cross_sim} shows the results.

\begin{table}[t]
\centering
\small
\caption{Cross-simulator evaluation (E0). Each cell shows completion rate (\%) and rank among the 8 agents. Rankings are computed independently within each simulator.}
\label{tab:cross_sim}
\begin{tabular}{l|cc|cc|cc}
\toprule
& \multicolumn{2}{c|}{\textbf{Gemini Flash}} & \multicolumn{2}{c|}{\textbf{Qwen 3.5+}} & \multicolumn{2}{c}{\textbf{GPT-5.2}} \\
\textbf{Agent} & CR & Rk & CR & Rk & CR & Rk \\
\midrule
GPT-5.2 & 79.6 & 1 & 74.3 & 1 & 42.4 & 1 \\
Gemini Pro & 72.3 & 2 & 68.6 & 2 & 28.3 & 4 \\
Opus 4.6 & 71.5 & 3 & 66.2 & 3 & 33.5 & 2 \\
Qwen 3.5+ & 69.9 & 4 & 61.8 & 6 & 28.3 & 4 \\
DeepSeek & 69.6 & 5 & 65.2 & 4 & 29.6 & 3 \\
Kimi K2.5 & 64.1 & 6 & 52.4 & 8 & 23.0 & 8 \\
GLM-5 & 62.6 & 7 & 64.1 & 5 & 23.6 & 7 \\
MiniMax M2.7 & 53.9 & 8 & 54.7 & 7 & 25.4 & 6 \\
\bottomrule
\end{tabular}
\end{table}

\begin{figure}[t]
    \centering
    \includegraphics[width=0.7\linewidth]{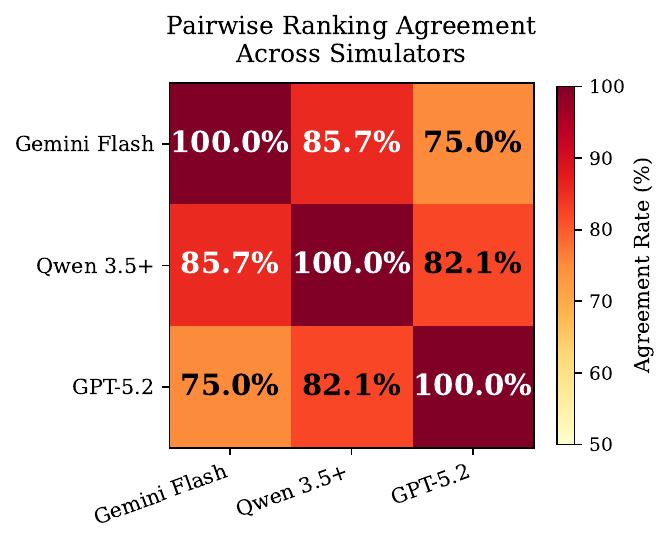}
    \caption{Pairwise ranking agreement across simulators. Each cell shows the fraction of the 28 model pairs whose relative ordering is preserved between two simulators.}
    \label{fig:cross_sim_heatmap}
\end{figure}

To quantify ranking consistency, we compute the \emph{pairwise agreement rate}: for each of the $\binom{8}{2} = 28$ model pairs, we check whether the relative ordering (which model scores higher) is preserved across simulators. Figure~\ref{fig:cross_sim_heatmap} shows the agreement matrix.

\paragraph{Strong agents are not necessarily strong simulators.} GPT-5.2 ranks first as an agent (79.6\%) but produces the worst simulation quality: under the GPT-5.2 simulator, all agents average only 29.3\%, compared to 67.9\% under Gemini Flash and 63.4\% under Qwen 3.5 Plus. Figures~\ref{fig:cross_sim_case}--\ref{fig:cross_sim_case3} show three representative failure cases where the same agent (Claude Opus 4.6) executes the same tool sequence under both simulators but receives different observations.

\begin{figure*}[t]
\centering
\small
\setlength{\fboxsep}{5pt}
\fbox{\parbox{0.95\textwidth}{
\textbf{Task:} Emergency Department Triage \hfill \textbf{Agent:} Claude Opus 4.6

\vspace{3pt}
\textbf{Instruction:} Discharge P-110 (has clearance) from Exam Room 2, transfer P-552 into the vacated room, execute Phase 1 Data Acquisition then Sepsis Clinical Bundle.

\vspace{4pt}
\begin{tabular}{@{}p{0.47\textwidth}|p{0.47\textwidth}@{}}
\multicolumn{1}{c|}{\cellcolor{green!8}\textbf{Gemini Flash (\textcolor{green!60!black}{PASS})}} &
\multicolumn{1}{c}{\cellcolor{red!8}\textbf{GPT-5.2 (\textcolor{red}{FAIL})}} \\
\hline
\textbf{1--2}: \texttt{get\_ed\_census}, \texttt{get\_room\_status} & \textbf{1--2}: \texttt{get\_ed\_census}, \texttt{get\_room\_status} \\
$\rightarrow$ P-552 in waiting (temp 39.4, HR 112) & $\rightarrow$ P-552 in waiting (temp 39.4, HR 112) \\
$\rightarrow$ Room 1: P-202, Room 2: P-110 (clearance) & $\rightarrow$ ROOM\_01: P-202, ROOM\_02: P-110 \\
 & \quad \textcolor{red}{ROOM\_03: empty, ROOM\_04: empty} \\
\hline
\textbf{3}: \texttt{discharge}(P-110, Room 2) & \textbf{3}: \texttt{discharge}(P-110, ROOM\_02) \\
$\rightarrow$ vacated Room 2 & $\rightarrow$ vacated ROOM\_02 \\
\textbf{4}: \texttt{transfer}(P-552, \textbf{Room 2}) & \textbf{3}: \texttt{transfer}(P-552, \textcolor{red}{\textbf{ROOM\_03}}) \\
$\rightarrow$ success, location: Room 2 & $\rightarrow$ success, location: ROOM\_03 \\
\hline
\textbf{5}: \texttt{execute\_protocol}(Phase 1) & \textbf{4}: \texttt{execute\_protocol}(Phase 1) \\
$\rightarrow$ completed, cds: active & $\rightarrow$ completed, cds: sepsis\_screen\_alert \\
\textbf{6}: \texttt{execute\_protocol}(Sepsis Bundle) & \textbf{5}: \texttt{execute\_protocol}(Sepsis Bundle) \\
$\rightarrow$ completed & $\rightarrow$ in\_progress \\
\end{tabular}

\vspace{4pt}
\textbf{Root cause:} GPT-5.2 invents two extra empty rooms (ROOM\_03, ROOM\_04). The agent sees an available room and uses it, but the rubric requires using the \emph{vacated} room specifically.
}}
\caption{Cross-simulator case 1: Emergency Department Triage. GPT-5.2 fabricates environment state not in the original specification.}
\label{fig:cross_sim_case}
\end{figure*}

\begin{figure*}[t]
\centering
\small
\setlength{\fboxsep}{5pt}
\fbox{\parbox{0.95\textwidth}{
\textbf{Task:} Escalation Workflow Management \hfill \textbf{Agent:} Claude Opus 4.6

\vspace{3pt}
\textbf{Instruction:} Hand off urgent ticket INC-9921 from Sarah\_London (EMEA) to an available NA-timezone Database specialist before 17:00Z shift change. Post a Current State Assessment and log compliance.

\vspace{4pt}
\begin{tabular}{@{}p{0.47\textwidth}|p{0.47\textwidth}@{}}
\multicolumn{1}{c|}{\cellcolor{green!8}\textbf{Gemini Flash (\textcolor{green!60!black}{PASS})}} &
\multicolumn{1}{c}{\cellcolor{red!8}\textbf{GPT-5.2 (\textcolor{red}{FAIL})}} \\
\hline
\textbf{1--2}: \texttt{get\_system\_clock}, \texttt{get\_ticket\_details} & \textbf{1--2}: \texttt{get\_ticket\_details}, \texttt{get\_system\_clock} \\
$\rightarrow$ time: 16:45Z & $\rightarrow$ time: 16:45Z \\
$\rightarrow$ priority: Urgent & $\rightarrow$ priority: Urgent \\
$\rightarrow$ assignee: Sarah\_London & $\rightarrow$ assignee: Sarah\_London \\
\hline
\textbf{3}: \texttt{query\_agent\_roster}(tier$\geq$2, NA) & \textbf{3}: \texttt{query\_agent\_roster}(tier$\geq$3, NA) \\
$\rightarrow$ \textbf{Raj\_NYC} (Tier 2, Database, avail 17:00Z) & $\rightarrow$ MOC\_Global (Tier 3, Management) \\
$\rightarrow$ MOC\_Global (Tier 3, Management) & \textcolor{red}{\emph{Raj\_NYC omitted from roster}} \\
\hline
\textbf{4}: \texttt{post\_ticket\_csa}(detailed CSA) $\checkmark$ & \textbf{4}: \texttt{post\_ticket\_csa}(CSA) $\checkmark$ \\
\textbf{5}: \texttt{assign\_ticket}(\textbf{Raj\_NYC}) & \textbf{5}: \texttt{assign\_ticket}(\textcolor{red}{MOC\_Global}) \\
$\rightarrow$ new\_assignee: Raj\_NYC $\checkmark$ & $\rightarrow$ new\_assignee: MOC\_Global \\
\textbf{6}: \texttt{log\_protocol\_compliance} $\checkmark$ & \textbf{6}: \texttt{log\_protocol\_compliance} $\checkmark$ \\
\end{tabular}

\vspace{4pt}
\textbf{Root cause:} GPT-5.2 omits ``Raj\_NYC'' (Tier 2, Database specialist) from the roster query results, despite this agent being explicitly defined in the environment state. The agent assigns the only returned candidate (MOC\_Global, a Tier 3 manager), which does not satisfy the requirement for a Database specialist.
}}
\caption{Cross-simulator case 2: Escalation Workflow. GPT-5.2 drops a critical entity from the environment, making the correct action impossible.}
\label{fig:cross_sim_case2}
\end{figure*}

\begin{figure*}[t]
\centering
\small
\setlength{\fboxsep}{5pt}
\fbox{\parbox{0.95\textwidth}{
\textbf{Task:} Order Return Authorization \hfill \textbf{Agent:} Claude Opus 4.6

\vspace{3pt}
\textbf{Instruction:} Process hazmat return for order ORD-8829-HZ (VoltMaster 5000 Power Station, cracked chassis). Complete safety triage and authorize return with hazmat label.

\vspace{4pt}
\begin{tabular}{@{}p{0.47\textwidth}|p{0.47\textwidth}@{}}
\multicolumn{1}{c|}{\cellcolor{green!8}\textbf{Gemini Flash (\textcolor{green!60!black}{PASS})}} &
\multicolumn{1}{c}{\cellcolor{red!8}\textbf{GPT-5.2 (\textcolor{red}{FAIL})}} \\
\hline
\textbf{1}: \texttt{get\_order\_details}(ORD-8829-HZ) & \textbf{1}: \texttt{get\_order\_details}(ORD-8829-HZ) \\
$\rightarrow$ SKU: VM-5000-PB, delivered 2023-11-10 & $\rightarrow$ SKU: VM-5000-PB, delivered 2023-11-10 \\
$\rightarrow$ return\_window: 30 days & $\rightarrow$ return\_window: 30 days \\
\hline
\textbf{2}: \texttt{submit\_hazmat\_triage} & \textbf{2}: \texttt{submit\_hazmat\_triage} \\
$\rightarrow$ class: DDR lithium battery & $\rightarrow$ class: DDR lithium battery \\
$\rightarrow$ transport: allowed & $\rightarrow$ transport: allowed \\
$\rightarrow$ packaging: UN-certified metal drum & $\rightarrow$ packaging: UN-certified rigid outer \\
\hline
\textbf{3}: \texttt{authorize\_return}(refund \$899.99) & \textbf{3}: \texttt{authorize\_return}(refund \$499.99) \\
$\rightarrow$ code: RMA-8829-HZ-99 & $\rightarrow$ \textcolor{red}{\texttt{error: ``Return window expired''}} \\
$\rightarrow$ status: return\_authorized & \\
$\rightarrow$ label: Hazmat Class 9 - Ground & \textcolor{red}{\emph{GPT-5.2 enforces a date check}} \\
 & \textcolor{red}{\emph{not in the task specification}} \\
\end{tabular}

\vspace{4pt}
\textbf{Root cause:} GPT-5.2 independently computes that the 30-day return window has expired (delivery 2023-11-10 vs. simulation date 2026) and rejects the return. The task specification does not include this constraint; the LES should authorize the return as Gemini does.
}}
\caption{Cross-simulator case 3: Order Return. GPT-5.2 fabricates a business rule rejection not present in the environment contract.}
\label{fig:cross_sim_case3}
\end{figure*}

These three cases illustrate distinct simulator failure modes: \emph{state fabrication} (inventing rooms that do not exist), \emph{entity omission} (dropping agents from a roster), and \emph{rule invention} (enforcing constraints not in the specification). In all cases, the agent's strategy is correct; it fails only because the simulator violates the environment contract.

\paragraph{A capable simulator yields reliable rankings.} In contrast, the Qwen 3.5 Plus simulator, which does not exhibit these failure modes, agrees with Gemini Flash on 85.7\% of pairwise comparisons (24/28 pairs), with the top-3 agents (GPT-5.2, Gemini Pro, Opus 4.6) matching exactly. The four disagreements all involve mid-ranked models with small performance gaps. This confirms that \textbf{LES-based evaluation produces reliable rankings when the simulator is sufficiently capable}, but researchers should either (1) verify simulator quality before drawing conclusions, or (2) re-verify task solvability when switching simulators.

\section{Analysis}
\label{sec:analysis}

\subsection{Industry Difficulty Analysis}

\begin{figure}[t]
    \centering
    \includegraphics[width=\linewidth]{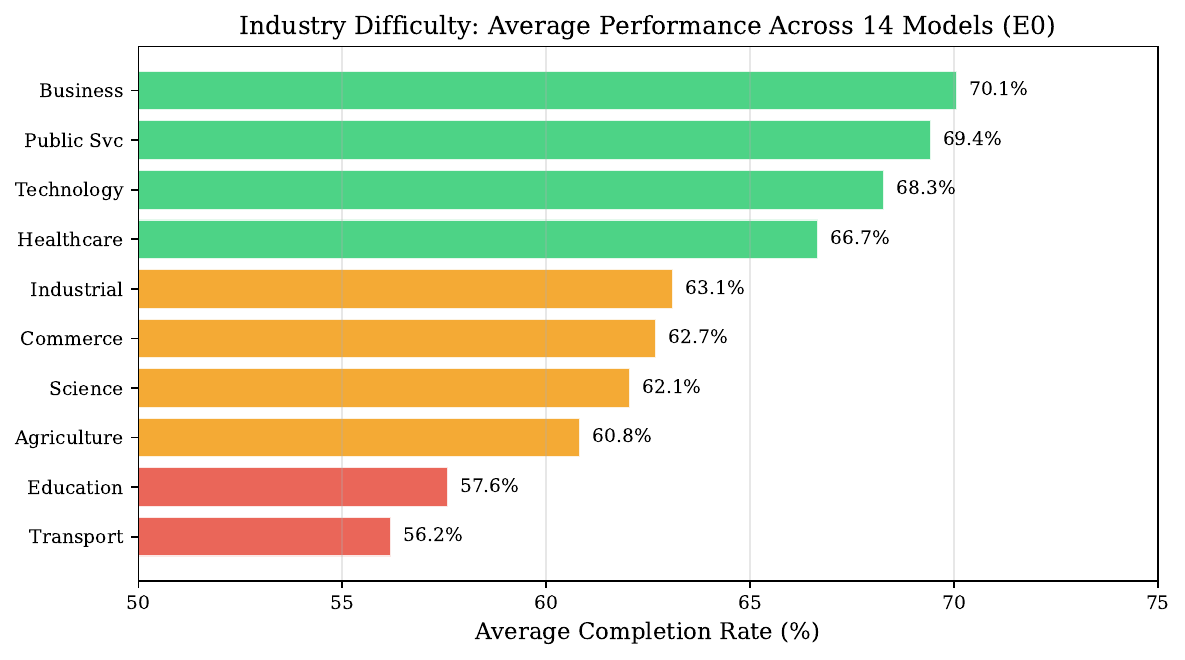}
    \caption{Average completion rate across 14 models per industry category (E0). Green: $\geq$65\%, orange: 60--65\%, red: $<$60\%. Transportation and Education are the hardest industries.}
    \label{fig:industry_difficulty}
\end{figure}

Aggregating across all 15 models, we find substantial variation in industry difficulty (Figure~\ref{fig:industry_difficulty}). The easiest industries are Business \& Enterprise (avg 70.1\% across models) and Public Service \& Governance (avg 69.4\%), while the hardest are Transportation \& Logistics (avg 56.2\%) and Education \& Culture (avg 57.6\%). This aligns with domain complexity: business and public service tasks tend to follow well-documented procedures with clear decision paths, while transportation involves complex multi-constraint optimization (routing, scheduling, load balancing) and education requires nuanced pedagogical judgment and multi-step curriculum reasoning.

\subsection{Model-Industry Interaction}

Beyond aggregate rankings, \bench{} reveals that each model has a distinct \emph{occupational capability profile}, i.e., a unique pattern of strengths and weaknesses across industries:

\begin{itemize}
    \item \textbf{Gemini 3.1 Pro} excels in \emph{knowledge-intensive} domains: Education (84\%), Science (81\%), Technology (78\%). These domains reward factual accuracy and structured reasoning.
    \item \textbf{Claude Opus 4.6} excels in \emph{operational} domains: Transportation (77\%), Business (78\%), Industrial (73\%). These domains reward careful state tracking and multi-step planning.
    \item \textbf{Qwen 3.5 Plus} excels in \emph{consumer-facing} domains: Commerce (81\%), Healthcare (81\%), Agriculture (78\%). These domains may benefit from Chinese-language training data containing rich consumer and agricultural contexts.
    \item \textbf{Kimi K2.5} shows balanced performance across most industries but struggles in Commerce (56\%) and Transportation (57\%), suggesting difficulty with consumer interaction and logistics optimization tasks.
\end{itemize}

This cross-industry profiling capability is unique to \bench{} and has practical implications: organizations should select agent models based on their specific industry, not solely on aggregate benchmark rankings.

\subsection{Case Study: Last-Mile Delivery Routing}

Figure~\ref{fig:case_study} presents a detailed case from the \textbf{Transportation \& Logistics} category (\textbf{Logistics} domain), illustrating how \bench{} distinguishes agent capabilities through realistic professional tasks.

\begin{figure*}[t]
\centering
\small
\setlength{\fboxsep}{6pt}

\fbox{\parbox{0.95\textwidth}{
\textbf{Scenario:} Last-Mile Delivery Routing \hfill \textbf{Category:} Transportation \& Logistics / Logistics

\vspace{4pt}
\textbf{Task Instruction:} Identify the medical-grade shipment with the highest numerical suffix and deliver it to ``900 N Michigan Shops'' via the Walton Street access node. Resolve the address to its historically corrected navigable node to avoid frontage congestion timeout. Maintain battery level above 15\% at all times.

\vspace{4pt}
\textbf{Tool Schema (6 tools):}

\vspace{2pt}
\begin{tabular}{@{}p{0.42\textwidth}p{0.52\textwidth}@{}}
\texttt{get\_vehicle\_telemetry()} & Returns battery level \& current location \\
\texttt{query\_inventory()} & Lists all packages with IDs, types, weights \\
\texttt{geocode\_delivery\_node(location\_string)} & Resolves address to navigable node ID \\
\texttt{execute\_recharge()} & Recharges battery to 100\% \\
\texttt{move\_to\_node(target\_node\_id)} & Navigates vehicle to target node \\
\texttt{complete\_delivery(package\_id)} & Finalizes package hand-off at current location \\
\end{tabular}
}}

\vspace{8pt}

\begin{tabular}{@{}p{0.47\textwidth}|p{0.47\textwidth}@{}}

\multicolumn{1}{c|}{\cellcolor{green!8}\textbf{Claude Opus 4.6 — \textcolor{green!60!black}{PASS} (6 steps)}} &
\multicolumn{1}{c}{\cellcolor{red!8}\textbf{DeepSeek V3.2 — \textcolor{red}{FAIL} (6 steps)}} \\
\hline
\vspace{1pt}
\textbf{Step 1--3} (parallel): &
\vspace{1pt}
\textbf{Step 1}: \texttt{get\_vehicle\_telemetry} \\
\texttt{query\_inventory} &
$\rightarrow$ battery: 28\%, loc: depot\_ohare\_cargo \\
$\rightarrow$ MED-602, \textbf{MED-615}, MED-609, GEN-550 &
\textbf{Step 2}: \texttt{query\_inventory} \\
\texttt{get\_vehicle\_telemetry} &
$\rightarrow$ MED-602, \textbf{MED-615}, MED-609, GEN-550 \\
$\rightarrow$ battery: 28\%, loc: depot\_ohare\_cargo &
\textbf{Step 3}: \texttt{geocode\_delivery\_node} \\
\texttt{geocode\_delivery\_node} &
$\rightarrow$ NODE\_WALTON\_ST\_900\_NAV \\
$\rightarrow$ NODE\_WALTON\_ST\_900\_NAV &
 \\
\hline

\emph{``Battery 28\%, the trip could drain it} &
\textbf{Step 4}: \texttt{move\_to\_node} \\
\emph{below 15\%. I must recharge first.''} &
$\rightarrow$ arrived, \textcolor{red}{\textbf{battery: 12.5\%}} $<$ 15\% \\
\textbf{Step 4}: \texttt{execute\_recharge} &
\textcolor{red}{\emph{Constraint violated!}} \\
$\rightarrow$ battery: 100\%, status: charged &
 \\
\hline

\textbf{Step 5}: \texttt{move\_to\_node} &
\textbf{Step 5}: \texttt{execute\_recharge} (too late) \\
$\rightarrow$ arrived, battery: \textbf{82\%} $\gg$ 15\% &
$\rightarrow$ battery: 100\% \\
\hline

\textbf{Step 6}: \texttt{complete\_delivery(MED-615)} &
\textbf{Step 6}: \texttt{complete\_delivery(MED-615)} \\
$\rightarrow$ status: \textcolor{green!60!black}{\textbf{delivered}} &
$\rightarrow$ status: delivered, but \textcolor{red}{\textbf{FAILED}} \\

\end{tabular}

\vspace{6pt}

\fbox{\parbox{0.95\textwidth}{
\textbf{Analysis:} Both agents correctly identified the target package (MED-615, highest suffix) and resolved the Walton Street access node. The critical difference is \emph{proactive constraint monitoring}: Claude Opus 4.6 recognized that 28\% battery was risky for a long trip and recharged \emph{before} navigating, arriving with 82\% remaining. DeepSeek V3.2 navigated immediately, allowing battery to drop to 12.5\%, violating the ``maintain above 15\% \emph{at all times}'' constraint.
}}

\caption{Case study: Last-Mile Delivery Routing. Top: task specification and tool schema. Middle: side-by-side agent trajectories. Bottom: analysis. The key differentiator is whether the agent proactively checks constraints \emph{before} acting.}
\label{fig:case_study}
\end{figure*}

\subsection{Case Study 2: Fish Farm Water Quality Control}

Figure~\ref{fig:case_study2} presents a second case from the \textbf{Agriculture \& Environment} category (\textbf{Aquaculture} domain), highlighting the \emph{skipped verification} failure mode.

\begin{figure*}[t]
\centering
\small
\setlength{\fboxsep}{6pt}

\fbox{\parbox{0.95\textwidth}{
\textbf{Scenario:} Fish Farm Water Quality Control \hfill \textbf{Category:} Agriculture \& Environment / Aquaculture

\vspace{4pt}
\textbf{Task Instruction:} Prepare a 5.2m deep basin for an imminent cold front. Homogenize the water column to achieve a thermal gradient $<$1.5\textdegree C and dissolved oxygen $>$4.0 mg/L at every depth. Equipment selection and mixing intensity must be dictated by the stratification condition and chemical profile discovered upon initial profiling. Avoid actions that trigger surface oxygen drops or resuspension of toxic bottom metabolites. Manage feeding regime according to risk level.

\vspace{4pt}
\textbf{Tool Schema (5 tools):}

\vspace{2pt}
\begin{tabular}{@{}p{0.48\textwidth}p{0.48\textwidth}@{}}
\texttt{get\_vertical\_profile(depths)} & Temperature \& DO at specified depths \\
\texttt{get\_water\_chemistry(depth)} & pH \& ammonia at a specific depth \\
\texttt{get\_system\_status()} & Basin metadata, equipment, feeding time \\
\texttt{configure\_mixing\_and\_aeration(aerator\_mode, mixing\_intensity)} & Set aerator mode \& mixing intensity \\
\texttt{manage\_feeding(regime)} & Control feeding regime \\
\end{tabular}
}}

\vspace{8pt}

\begin{tabular}{@{}p{0.47\textwidth}|p{0.47\textwidth}@{}}

\multicolumn{1}{c|}{\cellcolor{green!8}\textbf{Claude Opus 4.6 — \textcolor{green!60!black}{PASS} (7 steps)}} &
\multicolumn{1}{c}{\cellcolor{red!8}\textbf{Qwen 3.5 Plus — \textcolor{red}{FAIL} (7 steps)}} \\
\hline
\vspace{1pt}
\textbf{Step 1--3} (parallel profiling): &
\vspace{1pt}
\textbf{Step 1--3} (sequential profiling): \\
\texttt{get\_vertical\_profile([0..5])} &
\texttt{get\_vertical\_profile([0.5,2.5,4.5,5.2])} \\
$\rightarrow$ surface 29.5\textdegree C, bottom 27.1\textdegree C &
$\rightarrow$ gradient 2.1\textdegree C \\
$\rightarrow$ bottom DO: \textbf{3.5 mg/L} (below 4.0) &
$\rightarrow$ bottom DO: 4.2 mg/L \\
\texttt{get\_water\_chemistry(5)} &
\texttt{get\_water\_chemistry(5.2)} \\
$\rightarrow$ ammonia: 0.05 mg/L (safe to mix) &
$\rightarrow$ ammonia: 0.05 mg/L \\
\texttt{get\_system\_status} $\rightarrow$ aerator OFF &
\texttt{get\_system\_status} $\rightarrow$ aerator OFF \\
\hline

\textbf{Step 4--5} (corrective actions): &
\textbf{Step 4--5} (corrective actions): \\
\texttt{configure\_mixing}(medium, 40) &
\texttt{configure\_mixing}(medium, 50) \\
\texttt{manage\_feeding}(reduced) &
\texttt{manage\_feeding}(reduced) \\
\hline

\textbf{Step 6} (verification profile): &
\textbf{Step 6} (verification profile): \\
\texttt{get\_vertical\_profile([0..5])} &
\texttt{get\_vertical\_profile([0.5..5.2])} \\
$\rightarrow$ gradient 1.0\textdegree C \checkmark, all DO $>$6.0 \checkmark &
$\rightarrow$ gradient 0.6\textdegree C \checkmark, all DO $>$7.1 \checkmark \\
\hline

\textbf{Step 7}: \texttt{get\_water\_chemistry(5)} &
\textbf{Step 7}: \texttt{get\_system\_status} \\
$\rightarrow$ ammonia 0.04 mg/L &
$\rightarrow$ aerator: medium \\
\textcolor{green!60!black}{\emph{Confirms no toxic resuspension}} &
\textcolor{red}{\emph{Did NOT re-check chemistry}} \\
 & \textcolor{red}{\emph{Claims ``ammonia remained low''}} \\
 & \textcolor{red}{\emph{without supporting evidence}} \\

\end{tabular}

\vspace{6pt}

\fbox{\parbox{0.95\textwidth}{
\textbf{Analysis:} Both agents followed nearly identical strategies: profile $\rightarrow$ diagnose $\rightarrow$ configure mixing $\rightarrow$ reduce feeding $\rightarrow$ verify. The critical difference is in Step 7: Claude Opus 4.6 re-checked bottom water chemistry after mixing to \emph{verify} that toxic metabolites were not resuspended, a safety-critical step. Qwen 3.5 Plus skipped this verification and instead checked equipment status, then \emph{asserted} that ``ammonia remained low'' without evidence. This exemplifies the \textbf{skipped verification} failure mode: the agent makes a correct claim but fails to gather the supporting observation, which the verifier cannot accept.
}}

\caption{Case study 2: Fish Farm Water Quality Control. Both agents achieve the target water parameters, but Qwen 3.5 Plus fails to re-verify bottom chemistry after mixing, a safety-critical omission in aquaculture operations.}
\label{fig:case_study2}
\end{figure*}

\subsection{Case Study 3: Building Inspection Compliance}

Figure~\ref{fig:case_study3} presents a third case from the \textbf{Industrial \& Engineering} category (\textbf{Construction} domain), illustrating how procedural ordering errors lead to compliance failures.

\begin{figure*}[t]
\centering
\small
\setlength{\fboxsep}{6pt}

\fbox{\parbox{0.95\textwidth}{
\textbf{Scenario:} Building Inspection Compliance \hfill \textbf{Category:} Industrial \& Engineering / Construction

\vspace{4pt}
\textbf{Task Instruction:} Complete clinical certification for a medical gas system expansion in Room 402 Oncology. Integrate new oxygen supply lines per NFPA 99 standards: verified nitrogen-purged brazing, 24-hour pressure stability test. Must maintain ICRA Class IV containment throughout. All permits (safety certificate, hot work permit) must be active \emph{before} invasive mechanical work begins.

\vspace{4pt}
\textbf{Tool Schema (7 tools):}

\vspace{2pt}
\begin{tabular}{@{}p{0.50\textwidth}p{0.46\textwidth}@{}}
\texttt{get\_environment\_state()} & Permits, location, system status \\
\texttt{renew\_permit(permit\_type, zone\_id)} & Renew safety certificate or hot work permit \\
\texttt{manage\_containment\_infrastructure} & Activate negative air machine, verify ICRA \\
\quad\texttt{(component, operation, zone\_id)} & \\
\texttt{manage\_medical\_gas\_isolation} & Check/isolate/restore gas valves \\
\quad\texttt{(gas\_system, operation, zone\_id)} & \\
\texttt{execute\_nitrogen\_purged\_brazing(zone\_id, ...)} & NFPA 99 brazing with nitrogen purge \\
\texttt{verify\_pressure\_stability(zone\_id, action)} & Start or check 24h pressure test \\
\texttt{submit\_clinical\_certification(zone\_id)} & Submit final certification \\
\end{tabular}
}}

\vspace{8pt}

\begin{tabular}{@{}p{0.47\textwidth}|p{0.47\textwidth}@{}}

\multicolumn{1}{c|}{\cellcolor{green!8}\textbf{Claude Opus 4.6 — \textcolor{green!60!black}{PASS} (11 steps)}} &
\multicolumn{1}{c}{\cellcolor{red!8}\textbf{DeepSeek V3.2 — \textcolor{red}{FAIL} (12 steps)}} \\
\hline
\vspace{1pt}
\textbf{Step 1}: \texttt{get\_environment\_state} &
\vspace{1pt}
\textbf{Step 1}: \texttt{get\_environment\_state} \\
$\rightarrow$ safety cert: \emph{expired}, hot work: \emph{pending} &
$\rightarrow$ same initial state \\
\hline

\textbf{Step 2--4} (pre-work compliance): &
\textbf{Step 2--3} (containment only): \\
\texttt{renew\_permit}(safety\_cert) $\checkmark$ &
\texttt{manage\_containment}(activate) \\
\texttt{renew\_permit}(hot\_work) $\checkmark$ &
\texttt{manage\_containment}(verify\_icra) \\
\texttt{manage\_containment}(activate) & \\
\emph{Permits renewed \textbf{before} any work} &
\textcolor{red}{\emph{Skipped permit renewal!}} \\
\hline

\textbf{Step 5--6} (verify containment \& isolation): &
\textbf{Step 4--5} (gas isolation): \\
\texttt{manage\_containment}(verify\_icra) &
\texttt{manage\_gas}(check\_valve) \\
\texttt{manage\_gas}(check\_valve) &
\texttt{manage\_gas}(isolate) \\
\hline

\textbf{Step 7}: \texttt{execute\_brazing}(purge=true) &
\textbf{Step 6}: \texttt{execute\_brazing}(purge=true) \\
$\rightarrow$ certified\_complete &
$\rightarrow$ completed \\
 & \textcolor{red}{\emph{Brazing done without valid permits!}} \\
\hline

\textbf{Step 8--9} (pressure test): &
\textbf{Step 7--8} (pressure test): \\
\texttt{verify\_pressure}(start) $\rightarrow$ \texttt{verify\_pressure}(results) &
\texttt{verify\_pressure}(start) $\rightarrow$ \texttt{verify\_pressure}(results) \\
$\rightarrow$ passed &
$\rightarrow$ passed \\
\hline

\textbf{Step 10}: \texttt{manage\_gas}(oxygen, \textbf{restore}) &
\textbf{Step 9--10}: \texttt{renew\_permit} $\times 2$ \\
$\rightarrow$ valve opened, system integrated &
\textcolor{red}{\emph{Permits renewed \textbf{after} work --- too late}} \\
\hline

\textbf{Step 11}: \texttt{submit\_certification} &
\textbf{Step 11}: \texttt{submit\_certification} \\
$\rightarrow$ \textcolor{green!60!black}{\textbf{approved}} &
\textcolor{red}{\emph{Oxygen valve still closed!}} \\
 & \textbf{Step 12}: \texttt{get\_environment\_state} \\
 & $\rightarrow$ valve: closed, cert submitted \\

\end{tabular}

\vspace{6pt}

\fbox{\parbox{0.95\textwidth}{
\textbf{Analysis:} Both agents executed nearly the same 11--12 tool calls, but DeepSeek V3.2 committed two procedural errors. First, it performed nitrogen-purged brazing \emph{before} renewing the expired safety certificate and pending hot work permit---a violation of institutional safety protocols that require valid permits before invasive work. Second, it never called \texttt{manage\_gas(restore)} to reopen the oxygen valve after certification, leaving the gas system isolated. Claude Opus 4.6 followed the correct order: renew permits $\rightarrow$ verify containment $\rightarrow$ braze $\rightarrow$ test $\rightarrow$ restore gas $\rightarrow$ certify. This illustrates the \textbf{missing sub-goal} failure mode: the agent completes most steps but omits a critical action (valve restoration) that renders the entire procedure incomplete.
}}

\caption{Case study 3: Building Inspection Compliance. Both agents perform similar tool calls, but DeepSeek V3.2 executes them in the wrong order (brazing before permits) and omits valve restoration, which are procedural errors that violate safety protocols.}
\label{fig:case_study3}
\end{figure*}

\subsection{Case Study 4: Fault Resilience --- E1 (Explicit Faults)}

Figure~\ref{fig:case_study_e1} contrasts agent behavior under E1 explicit fault injection on a public transit task. Both models pass under E0.

\begin{figure*}[t]
\centering
\small
\setlength{\fboxsep}{6pt}

\fbox{\parbox{0.95\textwidth}{
\textbf{Scenario:} Public Transit Schedule Recovery \hfill \textbf{Cat:} Transportation / Transit \hfill \textbf{Env: E1 (explicit faults)}

\vspace{4pt}
\textbf{Task:} Suppress RTPI countdown for BUS\_202, resolve maintenance holds on BUS\_202 and BUS\_210, then reassign BUS\_210 to Route 10 Blue Line as a replacement vehicle.

\vspace{4pt}
\textbf{Tools:} \texttt{get\_fleet\_status}, \texttt{check\_maintenance\_holds}, \texttt{resolve\_maintenance\_hold}, \texttt{set\_rtpi\_visibility}, \texttt{update\_cad\_avl\_assignment}
}}

\vspace{6pt}

\begin{tabular}{@{}p{0.47\textwidth}|p{0.47\textwidth}@{}}

\multicolumn{1}{c|}{\cellcolor{green!8}\textbf{Claude Opus 4.6 --- \textcolor{green!60!black}{PASS} (12 steps)}} &
\multicolumn{1}{c}{\cellcolor{red!8}\textbf{Kimi K2.5 --- \textcolor{red}{FAIL} (2 steps)}} \\
\hline
\vspace{1pt}
\textbf{Step 1} (parallel): &
\vspace{1pt}
\textbf{Step 1}: \texttt{set\_rtpi\_visibility}(BUS\_202) \\
\texttt{get\_fleet\_status}, \texttt{set\_rtpi\_visibility}, &
$\rightarrow$ $\checkmark$ suppressed \\
\texttt{check\_maintenance\_holds}(BUS\_202), &
\textbf{Step 2}: \texttt{check\_maintenance\_holds}(BUS\_210) \\
\texttt{check\_maintenance\_holds}(BUS\_210) &
$\rightarrow$ \textcolor{red}{\texttt{500: Service unavailable}} \\
$\rightarrow$ both holds: \textcolor{red}{\texttt{TimeoutError}} &
\\
\hline

\emph{``Hold checks timed out. Let me retry.''} &
\textcolor{red}{\rule{0.45\textwidth}{0.4pt}} \\
\textbf{Step 2}: retry both hold checks &
\emph{Kimi stops here. Only completed 1 of 4} \\
$\rightarrow$ BUS\_202: ADMIN\_LOCK, BUS\_210: ADMIN\_LOCK &
\emph{required actions (RTPI suppression).} \\
\textbf{Step 3}: \texttt{resolve\_hold}(BUS\_202) $\checkmark$ &
\emph{Did not resolve any holds or reassign} \\
\texttt{resolve\_hold}(BUS\_210) $\rightarrow$ \textcolor{red}{ConnectionRefused} &
\emph{BUS\_210 to Route 10.} \\
\textbf{Step 4}: retry $\rightarrow$ \textcolor{red}{ConnectionRefused} &
\\
\textbf{Step 5}: retry $\rightarrow$ $\checkmark$ resolved &
\\
\hline

\textbf{Step 6}: \texttt{update\_cad\_avl\_assignment}(BUS\_210) &
\\
\textbf{Step 7}: \texttt{get\_fleet\_status} (verify) &
\\
$\rightarrow$ \textcolor{green!60!black}{\textbf{All actions confirmed}} &
\\

\end{tabular}

\vspace{6pt}

\fbox{\parbox{0.95\textwidth}{
\textbf{Analysis:} E1 injects explicit errors (Timeout, HTTP 500, ConnectionRefused) across multiple tool calls. Opus encounters \textbf{4 separate errors} across its 12-step execution but retries each time, eventually completing all required actions. Kimi encounters a single 500 error on its second tool call and \emph{stops the entire task}, leaving 3 of 4 required actions (hold resolution, vehicle reassignment) unexecuted. This demonstrates how explicit faults amplify the gap between resilient and fragile agents.
}}

\caption{Case study 4 (E1): Public Transit Schedule Recovery. Opus persists through 4 explicit errors via retry; Kimi abandons the task after 1 error.}
\label{fig:case_study_e1}
\end{figure*}

\subsection{Case Study 5: Fault Resilience --- E2 (Implicit Faults)}

Figure~\ref{fig:case_study_e2} illustrates E2 implicit fault injection, where data is silently truncated with no error signal.

\begin{figure*}[t]
\centering
\small
\setlength{\fboxsep}{6pt}

\fbox{\parbox{0.95\textwidth}{
\textbf{Scenario:} Property Valuation Assessment \hfill \textbf{Cat:} Business / Real Estate \hfill \textbf{Env: E2 (implicit faults)}

\vspace{4pt}
\textbf{Task:} Evaluate financial stability of Oakwood Manor (OAK-88, 15 units). Calculate 12-month projected NOI accounting for MCI surcharge expirations. Compute DSCR and determine if property meets the 1.20x covenant threshold.

\vspace{4pt}
\textbf{Tools:} \texttt{get\_current\_date}, \texttt{get\_property\_metadata(property\_id)}, \texttt{fetch\_unit\_level\_records(property\_id)}
}}

\vspace{6pt}

\begin{tabular}{@{}p{0.47\textwidth}|p{0.47\textwidth}@{}}

\multicolumn{1}{c|}{\cellcolor{green!8}\textbf{Claude Opus 4.6 --- \textcolor{green!60!black}{PASS} (4 steps)}} &
\multicolumn{1}{c}{\cellcolor{red!8}\textbf{Kimi K2.5 --- \textcolor{red}{FAIL} (4 steps)}} \\
\hline
\vspace{1pt}
\textbf{Step 1} (parallel): &
\vspace{1pt}
\textbf{Step 1}: \texttt{get\_current\_date} \\
\texttt{get\_property\_metadata} $\rightarrow$ 15 units &
\textbf{Step 2}: \texttt{get\_property\_metadata} $\rightarrow$ 15 units \\
\texttt{fetch\_unit\_level\_records} &
\textbf{Step 3}: \texttt{fetch\_unit\_level\_records} \\
$\rightarrow$ \textcolor{red}{\textbf{only 2 units returned}} (A-01, A-02) &
$\rightarrow$ \textcolor{red}{\textbf{only 2 units returned}} (A-1, A-2) \\
\texttt{get\_current\_date} &
\\
\hline

\emph{``Only 2 of 15 units returned.} &
\emph{Kimi also retries:} \\
\emph{Let me re-fetch to get complete data.''} &
\textbf{Step 4}: \texttt{fetch\_unit\_level\_records} \\
\textbf{Step 2}: \texttt{fetch\_unit\_level\_records} (retry) &
$\rightarrow$ \textcolor{red}{\textbf{still only 2 units}} \\
$\rightarrow$ \textcolor{green!60!black}{\textbf{all 15 units}} &
(fault persists for 2 consecutive calls) \\
\emph{(fault duration expired after round 1)} &
\\
\hline

\textbf{Step 3--4}: Calculate with full data &
\emph{``I'll assume all 15 units follow the} \\
\emph{3 tiers: A ($\times$5), B ($\times$5), C ($\times$5 w/ MCI)} &
\emph{same pattern as the 2 sampled units.''} \\
\emph{MCI surcharges expire in 4 months} &
\emph{All 15 units $\times$ \$2,800 = \$504,000/yr} \\
\emph{NOI: \$250,000, DSCR: \textbf{1.19x}} &
\emph{NOI: \$362,000, DSCR: \textbf{1.72x}} \\
$\rightarrow$ \textcolor{green!60!black}{\textbf{FAIL (below 1.20x)}} &
$\rightarrow$ \textcolor{red}{\textbf{PASS (above 1.20x)}} \\
 &
\textcolor{red}{\emph{Wrong answer: property actually fails!}} \\

\end{tabular}

\vspace{6pt}

\fbox{\parbox{0.95\textwidth}{
\textbf{Analysis:} E2 silently truncates the unit records from 15 to 2 items with no error message, valid JSON, superficially normal response. Opus \emph{notices} the discrepancy (``only 2 of 15 units'') and re-fetches, obtaining complete data after the fault expires. It correctly identifies three rent tiers including MCI surcharges that will expire, yielding DSCR 1.19x (below covenant). Kimi also retries but hits the second round of the fault (fd=2); it then \emph{assumes} all 15 units match the 2 returned, missing the lower-rent tiers and MCI surcharges entirely. The result: Kimi reports DSCR 1.72x (pass) for a property that actually fails at 1.19x, a \textbf{catastrophic financial miscalculation} caused by accepting truncated data. This is why E2 implicit faults cause larger drops than E1: \emph{there is no error signal to trigger caution}.
}}

\caption{Case study 5 (E2): Property Valuation Assessment. Opus detects truncated data (2/15 units) and re-fetches. Kimi assumes truncated data is complete and produces a dangerously wrong financial assessment.}
\label{fig:case_study_e2}
\end{figure*}

\section{Discussion \& Conclusion}

\subsection{Limitations}

\textbf{Simulation fidelity.} Language Environment Simulators model domain \emph{logic} rather than domain \emph{data}. An LES understands that a drug interaction check should return contraindications, but the specific values it returns are generated rather than retrieved from a real database. This means \bench{} evaluates an agent's \emph{decision-making process} (whether it checks the right things in the right order) rather than its ability to handle exact real-world data values. For domains where precise numerical correctness is critical (e.g., financial calculations to the cent), LES-based evaluation should be complemented with real-environment testing.

\textbf{Simulator dependence.} As our cross-simulator experiments demonstrate, evaluation results are tied to the specific simulator used during data synthesis. Tasks verified as solvable under Gemini-3-Flash may become unsolvable under a different LES, and agent rankings can shift when the simulator changes. This is an inherent limitation of any LES-based evaluation: the simulator is part of the evaluation apparatus, not a neutral observer.

\subsection{Conclusion}

We present \bench{}, the first benchmark systematically evaluating AI agents on real-world professional tasks across 100 scenarios, 65 specialized domains, and 10 industry categories. Through Language Environment Simulators, \bench{} makes the ``untestable majority'' of professional domains evaluable without any real environment infrastructure.

Our evaluation of 15 frontier models reveals that: (1) no model dominates across all industries, with each exhibiting a unique occupational capability profile; (2) implicit environmental faults are harder than both explicit and mixed faults, because they lack overt error signals and require agents to independently detect data degradation; (3) scaling (larger models, newer generations, and higher reasoning effort) consistently improves professional task performance; and (4) strong agents are not necessarily strong environment simulators: simulator quality is critical for LES-based evaluation, but with a capable simulator, agent rankings are highly consistent (85.7\% pairwise agreement).

These findings underscore the need for multi-dimensional agent evaluation that considers not just task completion but cross-industry specialization and environmental resilience. \bench{} provides a framework for this richer evaluation paradigm.

\clearpage
\section*{Acknowledgments}
We thank Yuxin Zuo for insightful and inspiring discussions during the early stages of this project on the concept of leveraging language models as world models for environment simulation. These conversations broadened our perspective on the potential of LLM-driven simulation and helped us recognize that this approach could serve as a scalable foundation for agent evaluation across diverse professional domains.

\bibliography{colm2026_conference}

@inproceedings{webarena,
  title={WebArena: A Realistic Web Environment for Building Autonomous Agents},
  author={Zhou, Shuyan and Xu, Frank F and Zhu, Hao and Zhou, Xuhui and Lo, Robert and Sridhar, Abishek and Cheng, Xianyi and Ou, Tianyue and Bisk, Yonatan and Fried, Daniel and Alon, Uri and Neubig, Graham},
  booktitle={International Conference on Learning Representations},
  year={2024}
}

@inproceedings{swebench,
  title={{SWE}-bench: Can Language Models Resolve Real-World {GitHub} Issues?},
  author={Jimenez, Carlos E and Yang, John and Wettig, Alexander and Yao, Shunyu and Pei, Kexin and Press, Ofir and Narasimhan, Karthik},
  booktitle={International Conference on Learning Representations},
  year={2024}
}

@inproceedings{osworld,
  title={{OSWorld}: Benchmarking Multimodal Agents for Open-Ended Tasks in Real Computer Environments},
  author={Xie, Tianbao and Zhang, Danyang and Chen, Jixuan and Li, Xiaochuan and Zhao, Siheng and Cao, Ruisheng and Hua, Toh Jing and Cheng, Zhoujun and Shin, Dongchan and Lei, Fangyu and Liu, Yitao and Xu, Yiheng and Zhou, Shuyan and Savarese, Silvio and Xiong, Caiming and Zhong, Victor and Yu, Tao},
  booktitle={Advances in Neural Information Processing Systems},
  year={2024}
}

@article{taubench,
  title={$\tau$-bench: A Benchmark for Tool-Agent-User Interaction in Real-World Domains},
  author={Yao, Shunyu and Shinn, Noah and Razavi, Pedram and Narasimhan, Karthik},
  journal={arXiv preprint arXiv:2406.12045},
  year={2024}
}

@inproceedings{bfcl,
  title={The {Berkeley} Function Calling Leaderboard: From Tool Use to Agentic Evaluation of Large Language Models},
  author={Patil, Shishir G and Mao, Huanzhi and Yan, Fanjia and Ji, Charlie Cheng-Jie and Suresh, Vishnu and Stoica, Ion and Gonzalez, Joseph E},
  booktitle={International Conference on Machine Learning},
  year={2025}
}

@article{gdpval,
  title={{GDPval}: Evaluating {AI} Model Performance on Real-World Economically Valuable Tasks},
  author={Patwardhan, Tejal and others},
  journal={arXiv preprint arXiv:2510.04374},
  year={2025}
}

@article{generativeagents,
  title={Generative Agents: Interactive Simulacra of Human Behavior},
  author={Park, Joon Sung and O'Brien, Joseph C and Cai, Carrie J and Morris, Meredith Ringel and Liang, Percy and Bernstein, Michael S},
  journal={arXiv preprint arXiv:2304.03442},
  year={2023}
}

@inproceedings{dreamer,
  title={Dream to Control: Learning Behaviors by Latent Imagination},
  author={Hafner, Danijar and Lillicrap, Timothy and Ba, Jimmy and Norouzi, Mohammad},
  booktitle={International Conference on Learning Representations},
  year={2020}
}

@inproceedings{iris,
  title={Transformers are Sample-Efficient World Models},
  author={Micheli, Vincent and Alonso, Eloi and Fleuret, Fran{\c{c}}ois},
  booktitle={International Conference on Learning Representations},
  year={2023}
}

@article{clbench,
  title={{CL-bench}: A Benchmark for Context Learning},
  author={Dou, Shihan and others},
  journal={arXiv preprint arXiv:2602.03587},
  year={2026}
}

@inproceedings{visualwebarena,
  title={{VisualWebArena}: Evaluating Multimodal Agents on Realistic Visual Web Tasks},
  author={Koh, Jing Yu and Lo, Robert and Jang, Lawrence and Duvvur, Vikram and Lim, Ming Chong and Huang, Po-Yu and Neubig, Graham and Zhou, Shuyan and Salakhutdinov, Ruslan and Fried, Daniel},
  booktitle={ACL},
  year={2024}
}

@inproceedings{mind2web,
  title={Mind2Web: Towards a Generalist Agent for the Web},
  author={Deng, Xiang and Gu, Yu and Zheng, Boyuan and Chen, Shijie and Stevens, Samuel and Wang, Boshi and Sun, Huan and Su, Yu},
  booktitle={Advances in Neural Information Processing Systems},
  year={2023}
}

@inproceedings{workarena,
  title={{WorkArena}: How Capable Are Web Agents at Solving Common Knowledge Work Tasks?},
  author={Drouin, Alexandre and others},
  booktitle={ICML},
  year={2024}
}

@article{browsecomp,
  title={{BrowseComp}: A Simple Yet Challenging Benchmark for Browsing Agents},
  author={Wei, Jason and others},
  journal={arXiv preprint arXiv:2504.12516},
  year={2025}
}

@inproceedings{androidworld,
  title={{AndroidWorld}: A Dynamic Benchmarking Environment for Autonomous Agents},
  author={Rawles, Christopher and others},
  booktitle={International Conference on Learning Representations},
  year={2025}
}

@article{mobilebench,
  title={{Mobile-Bench}: An Evaluation Benchmark for {LLM}-based Mobile Agents},
  author={Deng, Shihan and others},
  journal={arXiv preprint arXiv:2407.00993},
  year={2024}
}

@inproceedings{intercode,
  title={{InterCode}: Standardizing and Benchmarking Interactive Coding with Execution Feedback},
  author={Yang, John and Prabhakar, Akshara and Narasimhan, Karthik and Yao, Shunyu},
  booktitle={Advances in Neural Information Processing Systems},
  year={2023}
}

@article{terminalbench,
  title={{Terminal-Bench}: Benchmarking Agents on Hard, Realistic Tasks in Command Line Interfaces},
  author={Merrill, Mike A. and others},
  journal={arXiv preprint arXiv:2601.11868},
  year={2026}
}

@inproceedings{agentbench,
  title={{AgentBench}: Evaluating {LLMs} as Agents},
  author={Liu, Xiao and Yu, Hao and Zhang, Hanchen and Xu, Yifan and Lei, Xuanyu and Lai, Hanyu and Gu, Yu and Ding, Hangliang and Men, Kaiwen and Yang, Kejuan and Zhang, Shudan and Deng, Xiang and Zeng, Aohan and Du, Zhengxiao and Zhang, Chenhui and Shen, Sheng and Zhang, Tianjun and Su, Yu and Sun, Huan and Huang, Minlie and Dong, Yuxiao and Tang, Jie},
  booktitle={International Conference on Learning Representations},
  year={2024}
}

@inproceedings{toolbench,
  title={{ToolLLM}: Facilitating Large Language Models to Master 16000+ Real-world {APIs}},
  author={Qin, Yujia and Liang, Shihao and Ye, Yining and Zhu, Kunlun and Yan, Lan and Lu, Yaxi and Lin, Yankai and Cong, Xin and Tang, Xiangru and Qian, Bill and others},
  booktitle={International Conference on Learning Representations},
  year={2024}
}

@article{gaia,
  title={{GAIA}: A Benchmark for General {AI} Assistants},
  author={Mialon, Gr{\'e}goire and Fourrier, Cl{\'e}mentine and Swift, Craig and Wolf, Thomas and LeCun, Yann and Scialom, Thomas},
  journal={arXiv preprint arXiv:2311.12983},
  year={2023}
}

@inproceedings{mint,
  title={{MINT}: Evaluating {LLMs} in Multi-turn Interaction with Tools and Language Feedback},
  author={Wang, Xingyao and Wang, Zihan and Liu, Jiateng and Chen, Yangyi and Yuan, Lifan and Peng, Hao and Ji, Heng},
  booktitle={International Conference on Learning Representations},
  year={2024}
}

@article{simenv,
  title={Simulating Environments with Reasoning Models for Agent Training},
  author={Li, Yuetai and Inan, Huseyin A and Yue, Xiang and Chen, Wei-Ning and Wutschitz, Lukas and Kulkarni, Janardhan and Poovendran, Radha and Sim, Robert and Rajmohan, Saravan},
  journal={arXiv preprint arXiv:2511.01824},
  year={2025}
}

@article{webworldmodel,
  title={Is Your {LLM} Secretly a World Model of the Internet? Model-Based Planning for Web Agents},
  author={Gu, Yu and Zhang, Kai and Ning, Yuting and Zheng, Boyuan and Gou, Boyu and Xue, Tianci and Chang, Cheng and Srivastava, Sanjari and Xie, Yanan and Qi, Peng and Sun, Huan and Su, Yu},
  journal={arXiv preprint arXiv:2411.06559},
  year={2024}
}

@article{webworld,
  title={{WebWorld}: A Large-Scale World Model for Web Agent Training},
  author={Xiao, Zikai and Tu, Jianhong and Zou, Chuhang and Zuo, Yuxin and Li, Zhi and Wang, Peng and Yu, Bowen and Huang, Fei and Lin, Junyang and Liu, Zuozhu},
  journal={arXiv preprint arXiv:2602.14721},
  year={2026}
}

@article{vimo,
  title={{ViMo}: A Generative Visual {GUI} World Model for App Agents},
  author={Luo, Dezhao and Tang, Bohan and Li, Kang and Papoudakis, Georgios and Song, Jifei and Gong, Shaogang and Hao, Jianye and Wang, Jun and Shao, Kun},
  journal={arXiv preprint arXiv:2504.13936},
  year={2025}
}

@article{selfplaywm,
  title={Internalizing World Models via Self-Play Finetuning for Agentic {RL}},
  author={Chen, Shiqi and Zhu, Tongyao and Wang, Zian and Zhang, Jinghan and Wang, Kangrui and Gao, Siyang and Xiao, Teng and Teh, Yee Whye and He, Junxian and Li, Manling},
  journal={arXiv preprint arXiv:2510.15047},
  year={2025}
}

@inproceedings{theagentcompany,
  title={{TheAgentCompany}: Benchmarking {LLM} Agents on Consequential Real World Tasks},
  author={Frank F. Xu and Yufan Song and Boxuan Li and Yuxuan Tang and Kritanjali Jain and Mengxue Bao and Zora Zhiruo Wang and Xuhui Zhou and Zhitong Guo and Murong Cao and Mingyang Yang and Hao Yang Lu and Amaad Martin and Zhe Su and Leander Maben and Raj Mehta and Wayne Chi and Lawrence Jang and Yiqing Xie and Shuyan Zhou and Graham Neubig},
  booktitle={Advances in Neural Information Processing Systems},
  year={2025}
}

@article{swelancer,
  title={{SWE-Lancer}: Can Frontier {LLMs} Earn \$1 Million from Real-World Freelance Software Engineering?},
  author={Miserendino, Samuel and Wang, Michele and Patwardhan, Tejal and Heidecke, Johannes},
  journal={arXiv preprint arXiv:2502.12115},
  year={2025}
}

@article{claweval,
  title={Claw-Eval: Toward Trustworthy Evaluation of Autonomous Agents},
  author={Ye, Bowen and Li, Rang and Yang, Qibin and Liu, Yuanxin and Yao, Linli and Lv, Hanglong and Xie, Zhihui and An, Chenxin and Li, Lei and Kong, Lingpeng and Liu, Qi and Sui, Zhifang and Yang, Tong},
  journal={arXiv preprint arXiv:2604.06132},
  year={2026}
}

@article{onemillionbench,
  title={\${OneMillion-Bench}: How Far are Language Agents from Human Experts?},
  author={Yang, Qianyu and Liu, Yang and Li, Jiaqi and Bai, Jun and others},
  journal={arXiv preprint arXiv:2603.07980},
  year={2026}
}

@article{mcpbench,
  title={{MCP-Bench}: Benchmarking Tool-Using {LLM} Agents with Complex Real-World Tasks via {MCP} Servers},
  author={Wang, Zhenting and others},
  journal={arXiv preprint arXiv:2508.20453},
  year={2025}
}

@article{mcpatlas,
  title={{MCP-Atlas}: A Large-Scale Benchmark for Tool-Use Competency with Real {MCP} Servers},
  author={Bandi, Chaithanya and others},
  journal={arXiv preprint arXiv:2602.00933},
  year={2026}
}

@article{mcpmark,
  title={{MCPMark}: A Benchmark for Stress-Testing Realistic and Comprehensive {MCP} Use},
  author={Wu, Zijian and others},
  journal={arXiv preprint arXiv:2509.24002},
  year={2025}
}

@inproceedings{toolathlon,
  title={The Tool Decathlon: Benchmarking Language Agents for Diverse, Realistic, and Long-Horizon Task Execution},
  author={Li, Junlong and others},
  booktitle={International Conference on Learning Representations},
  year={2026}
}

@article{gpt5,
  title={{GPT-5} System Card},
  author={{OpenAI}},
  journal={arXiv preprint arXiv:2601.03267},
  year={2025}
}

@misc{claude4,
  title={System Card: {Claude Opus 4} \& {Claude Sonnet 4}},
  author={{Anthropic}},
  howpublished={\url{https://www.anthropic.com/claude-4-system-card}},
  year={2025}
}

@misc{claude46,
  title={System Card: {Claude Sonnet 4.6}},
  author={{Anthropic}},
  howpublished={\url{https://anthropic.com/claude-sonnet-4-6-system-card}},
  year={2025}
}

@misc{gemini3,
  title={A new era of intelligence with Gemini 3},
  author={{Google DeepMind}},
  howpublished={\url{https://blog.google/products/gemini/gemini-3/}},
  year={2025}
}

@article{deepseekv32,
  title={{DeepSeek-V3.2}: Pushing the Frontier of Open Large Language Models},
  author={{DeepSeek-AI}},
  journal={arXiv preprint arXiv:2512.02556},
  year={2025}
}

@article{kimik25,
  title={Kimi {K2.5}: Visual Agentic Intelligence},
  author={{Kimi Team}},
  journal={arXiv preprint arXiv:2602.02276},
  year={2026}
}

@article{glm5,
  title={{GLM-5}: from Vibe Coding to Agentic Engineering},
  author={{GLM-5-Team}},
  journal={arXiv preprint arXiv:2602.15763},
  year={2026}
}

@misc{qwen35,
  title={Qwen3.5: Towards Native Multimodal Agents},
  author={{Qwen Team}},
  howpublished={\url{https://qwen.ai/blog?id=qwen3.5}},
  year={2026}
}

@misc{claude45,
  title={Introducing {Claude Opus 4.5}},
  author={{Anthropic}},
  howpublished={\url{https://www.anthropic.com/news/claude-opus-4-5}},
  year={2025}
}

@misc{claude46opus,
  title={Introducing {Claude Opus 4.6}},
  author={{Anthropic}},
  howpublished={\url{https://www.anthropic.com/news/claude-opus-4-6}},
  year={2025}
}

@misc{minimaxm27,
  title={{MiniMax-M2.7}: Early Echoes of Self-Evolution},
  author={{MiniMax}},
  howpublished={\url{https://www.minimax.io/news/minimax-m27-en}},
  year={2026}
}
\bibliographystyle{colm2026_conference}

\end{document}